\PassOptionsToPackage{svgnames}{xcolor}
\documentclass[journal]{vgtc}  

\vgtccategory{Research}

\title{VIPA: View-Invariant Projector-Based Adversarial Attack}

\author{
  \authororcid{Jiyu Han}{0009-0006-4032-3578},
  \authororcid{Qingyue Deng}{0009-0001-8471-2480}, and
  \authororcid{Bingyao Huang}{0000-0002-8647-5730}
}

\authorfooter{
  \item
  	Jiyu Han is with Southwest University.
  	E-mail: hjy202308@email.swu.edu.cn.
  \item
  	Qingyue Deng is with Southwest University.
  	E-mail: qingyue@email.swu.edu.cn.

  \item Bingyao Huang is with Southwest University.
  	E-mail: bhuang@swu.edu.cn. Corresponding author.
}

\abstract{
Projector-based adversarial attack aims to physically manipulate real-world scenes by projecting adversarial patterns, thereby causing deep image classifiers to produce incorrect predictions. However, existing stealthy projector-based adversarial attack methods model the project-and-capture process in a single and static view, limiting their view-invariant capability and practical applicability in dynamic environments. In this paper, we introduce View-Invariant Projector-Based Adversarial Attack (VIPA), a novel method designed to overcome these limitations by achieving both view-invariant and classifier-agnostic adversarial attacks in the physical world. We first leverage a view-invariant projector-camera system simulation method to model the physical interactions between projected patterns and real-world surfaces. To ensure robustness and stealthiness of the attack across different views, VIPA optimizes adversarial projections by aggregating simulated attack losses from multiple views. This joint optimization across diverse views helps maintain robustness regardless of the viewpoints. Finally, the optimized patterns are physically projected into real-world scenes, where they successfully fool various classifiers from different viewpoints, thereby enabling robust and practical view-invariant adversarial attacks. Our experiments in both targeted and untargeted attacks demonstrate that VIPA consistently achieves higher attack success rates than existing methods, while also enhancing stealthiness and ensuring minimal perceptual degradation across all tested views. Please refer to our project page for the code and dataset: \url{https://Hhhh-han.github.io/VIPA/}.
} 


\teaser{
  \centering
  \includegraphics[width=0.96\linewidth]{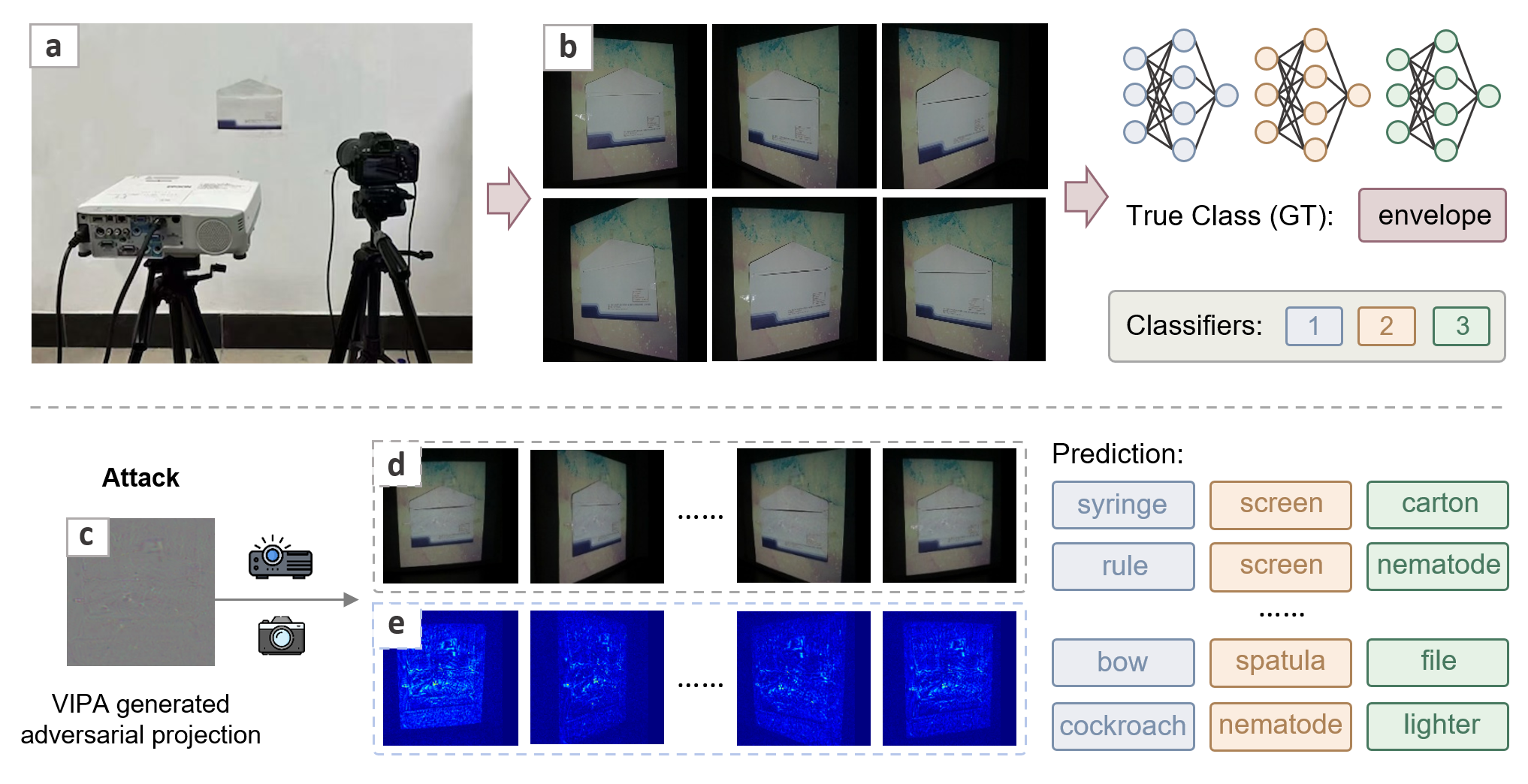}
  \caption{View-Invariant Projector-Based Adversarial Attack (VIPA): \textbf{(a)} System setup: consists of a projector, a camera, and a physical object. 
  \textbf{(b)} Scene images captured from different viewpoints by the camera under normal lighting conditions. These images are classified by different classifiers, which unanimously predict the object as \textbf{envelope}.  
  \textbf{(c)} An adversarial projection generated by our VIPA, designed such that when projected onto the object, the resulting camera-captured scene can be misclassified.  
  \textbf{(d)} Camera-captured adversarial scenes from unseen viewpoints. The corresponding predictions from three classifiers are shown on the right, where each row represents one viewpoint.
  \textbf{(e)} Normalized difference maps between (d) and their corresponding clean scenes (analogous to (b)) captured from the same viewpoints.
  Clearly, the camera-captured adversarial projection remains visually stealthy while consistently fooling different classifiers from multiple viewpoints. More results are provided in \autoref{sec:experiments} and supplementary.}
  \label{fig:teaser}
}

\graphicspath{{figures/}{pictures/}{images/}{./}} 

\usepackage{times}

\usepackage{tabu}                     
\usepackage{booktabs}                
\usepackage{lipsum}                  
\usepackage{mwe}                     
\usepackage[ruled,vlined]{algorithm2e}
\usepackage{algorithmic}
\usepackage{amsmath,amssymb}
\usepackage{amsfonts}
\usepackage{mathtools}

\usepackage{multirow}
\usepackage{cite}
\usepackage{textcomp}   
\usepackage{xcolor} 
\usepackage{colortbl}
\usepackage{pifont}
\usepackage{tikz}

\AtBeginDocument{
\renewcommand{\equationautorefname}{Eqn.}
\renewcommand{\sectionautorefname}{\S}
\renewcommand{\subsectionautorefname}{\S}

}

\definecolor{green}{RGB}{152, 194, 130}
\definecolor{red}{RGB}{230, 117, 101}
\definecolor{white}{RGB}{255, 255, 255}
\definecolor{gray}{RGB}{230,230,230}

\definecolor{softgreen}{HTML}{007A6E}
\definecolor{softred}{HTML}{CF2923}

\definecolor{cblue_border}{HTML}{7B90AB}
\definecolor{cblue_fill}{HTML}{EBEDF2}
\definecolor{corange_border}{HTML}{AD8358}
\definecolor{corange_fill}{HTML}{FBEDE0}
\definecolor{cgreen_border}{HTML}{487960}
\definecolor{cgreen_fill}{HTML}{E6F2E6}

\newcommand{\colorboxblock}[2]{%
    \begin{tikzpicture}[baseline=-0.3em]
        \node[
            rectangle,
            draw=#1,      
            fill=#2,      
            rounded corners=1.5pt,
            line width=1pt,       
            minimum width=1.2em,  
            minimum height=0.8em,  
            inner sep=0pt
        ] {};
    \end{tikzpicture}
}

\newcommand{\cmark}{\textcolor{softgreen}{\ding{51}}} 
\newcommand{\xmark}{\textcolor{softred}{\ding{55}}}  

\newlength{\myboxsep}
\newcommand{\supptitle}[1]{\def\mySuppTitleContent{#1}}
\newcommand{\makeMySupplementaryTitle}{%
  \onecolumn
  \begin{center}
    {\sffamily\huge
    \mySuppTitleContent\par}
    \vspace{8pt}
    {\large\sffamily\itshape
    \textemdash\quad Supplementary Materials\quad\textemdash\par}
    \vspace{14pt}
    {\large\sffamily
    \authororcid{Jiyu Han}{0009-0006-4032-3578},
    \authororcid{Qingyue Deng}{0009-0001-8471-2480}, and
    \authororcid{Bingyao Huang}{0000-0002-8647-5730}\par}
    \vspace{2em}
  \end{center}
}

\begin{document}

\firstsection{Introduction}

\maketitle

\begin{table*}[!t]
    \centering
    \caption{
    Comparison of our proposed VIPA against existing projector-based adversarial attack methods. The table contrasts these approaches along five aspects: the stealthiness level and visual appearance of the projected perturbations, whether the attack is \textbf{classifier-agnostic (CA)}, whether it is \textbf{view-invariant (VI)}, and the ProCams simulation used to model the project-and-capture process. Early methods \cite{nichols2018projectingtroublelightbased, OPAD2021, hu2023adversarial} often introduce noticeable artifacts such as isolated pixels/squares, structured patterns, or global color casts, resulting in limited stealthiness. SPAA \cite{SPAA2022} moves toward much stealthier perturbations through noise-like projections, but its Convolutional Neural Network based (CNN-based) simulation reduces robustness to viewpoint changes. CAPAA \cite{CAPAA2025} further improves imperceptibility through attention-based perturbations and attains classifier-agnostic attacks, yet it is not view-invariant. In contrast, VIPA leverages Gaussian Splatting with multiple views losses to enable view-invariant attacks.
    }
    \label{tab:comparison}
    \resizebox{\textwidth}{!}{
        \fontsize{5pt}{7pt}\selectfont
        \begin{tabular}{rccccc}
            \toprule
            Method & Visual appearance & Stealthiness level & CA & VI & ProCams simulation \\
            \midrule
            One-pixel DE \cite{nichols2018projectingtroublelightbased} & Obvious pixels/squares & Low & \xmark & \xmark & None \\
            OPAD \cite{OPAD2021} & Structured patterns & Medium & \xmark & \xmark & Explicit radiometric model \\
            AdvCP \cite{hu2023adversarial} & Global color cast & Medium & \xmark & \xmark & Linear fusion + EOT \\
            SPAA \cite{SPAA2022} & Noise & High & \xmark & \xmark & CNN \\
            CAPAA \cite{CAPAA2025} & Attention-based noise & High & \cmark & \xmark & CNN \\
            VIPA (ours) & View-invariant noise & High & \cmark & \cmark & Gaussian splatting \\
            \bottomrule
        \end{tabular}
    }
\end{table*}

Spatial Augmented Reality (SAR) enhances human interaction with the physical world by embedding digital content directly onto real surfaces, most commonly through projector–camera (ProCam) systems \cite{kim2023reality, SAR_industrial, SAR_car}. Compared to head-mounted displays, SAR eliminates wearable constraints and enables shared, room-scale experiences, but also imposes requirements on geometric calibration and radiometric compensation to maintain visual fidelity on complex, textured, and non-Lambertian surfaces. As SAR systems increasingly evolve from passive visualization to interactive and autonomous environments, they rely heavily on deep neural networks (DNNs) for real-time perception tasks such as object recognition, detection, and pose estimation \cite{Deep2025}.

However, the reliance on deep learning also introduces a critical vulnerability to adversarial examples. While DNNs have achieved remarkable success in interpreting complex visual scenes, they remain highly vulnerable to adversarial attacks, which involve subtle and carefully crafted perturbations that mislead models into producing incorrect predictions. In augmented and mixed reality settings, such vulnerabilities raise security concerns because perception errors can propagate to downstream decision-making and potentially affect safety-critical behaviors \cite{ISMAR25bst}. Projector-based adversarial attacks, in particular, manipulate the visual appearance of physical objects by projecting carefully crafted light patterns onto real-world scenes. In the context of SAR, this technique introduces a distinct threat vector because, unlike conventional digital attacks \cite{carlini2017, dong2018, zhao2020largeimperceptibleadversarialimage, su2019one, rony2019decouplingdirectionnormefficient, moosavidezfooli2017universaladversarialperturbations} that directly perturb image pixels, it exploits the transient and dynamic nature of light to enable real-time modifications without permanently altering the physical environment. This capability mirrors SAR’s programmable appearance but can be leveraged to expose the fragility of modern recognition systems. Despite its potential, designing effective projector-based attacks remains challenging, as the projected perturbations must maintain robustness across diverse viewpoints and multiple image classifiers while simultaneously preserving sufficient visual stealthiness to evade human detection.

Existing projector-based adversarial attack methods that employ neural networks \cite{SPAA2022, CAPAA2025} have made notable progress in enhancing the stealthiness of adversarial projections, improving success rates for specific viewpoints, and achieving classifier-agnostic performance, yet they still lack view-invariant capability, which limits their effectiveness in realistic scenarios where objects are observed from varying viewpoints, such as dynamic scenes or multi-camera environments. This limitation arises because both SPAA \cite{SPAA2022} and CAPAA \cite{CAPAA2025} rely on PCNet, a convolutional neural network-based ProCams simulation framework, to model the physical projection process and its interaction with captured images. While this enables differentiable optimization of adversarial patterns for specific camera poses, it inherently binds the resulting projections to the sampled or simulated views.

In this paper, we propose VIPA, the first projector-based adversarial attack method that achieves both view-invariant and classifier-agnostic capabilities simultaneously, thereby overcoming these limitations. Our VIPA consists of two main components: 
(1) a projector–camera simulation process, which uses 2D Gaussian modeling to represent the scene, and 
(2) an attack optimization process that generates the adversarial projection. To achieve view-invariant attacks, VIPA introduces a novel multi-view loss aggregation strategy that effectively integrates the losses from different views during optimization. This approach enables the generation of a single adversarial projection that remains effective under varying camera angles while generalizing across different classifiers without requiring model-specific tuning. As a result, VIPA improves the attack success rate for randomly selected views and enhances the stealthiness of the projection in realistic physical environments.

Our contributions can be summarized as follows:
\begin{itemize}
   
    \item The proposed VIPA is the first projector-based adversarial attack method that is both view-invariant and classifier-agnostic.
    \item We design a novel per-view loss aggregation strategy to enable robustness against variations in camera viewpoints.
    \item Experimental comparisons clearly show that our VIPA outperforms previous method in both attack success rate and stealthiness.
\end{itemize}

In the rest of the paper, we introduce the related work in \autoref{sec:related_work}, and describe the problem formulation and the proposed VIPA in \autoref{sec:methods}. We show our system configurations and experimental evaluations in \autoref{sec:experiments}, discuss the paper in \autoref{sec:discussion}, and conclude the paper in \autoref{sec:conclusion}.

\section{Related Work}\label{sec:related_work}

In this section, we review prior work primarily in two key areas,  adversarial attack and ProCams simulation.

\begin{figure*}[!t]
\begin{center}
\includegraphics[width=1\linewidth]{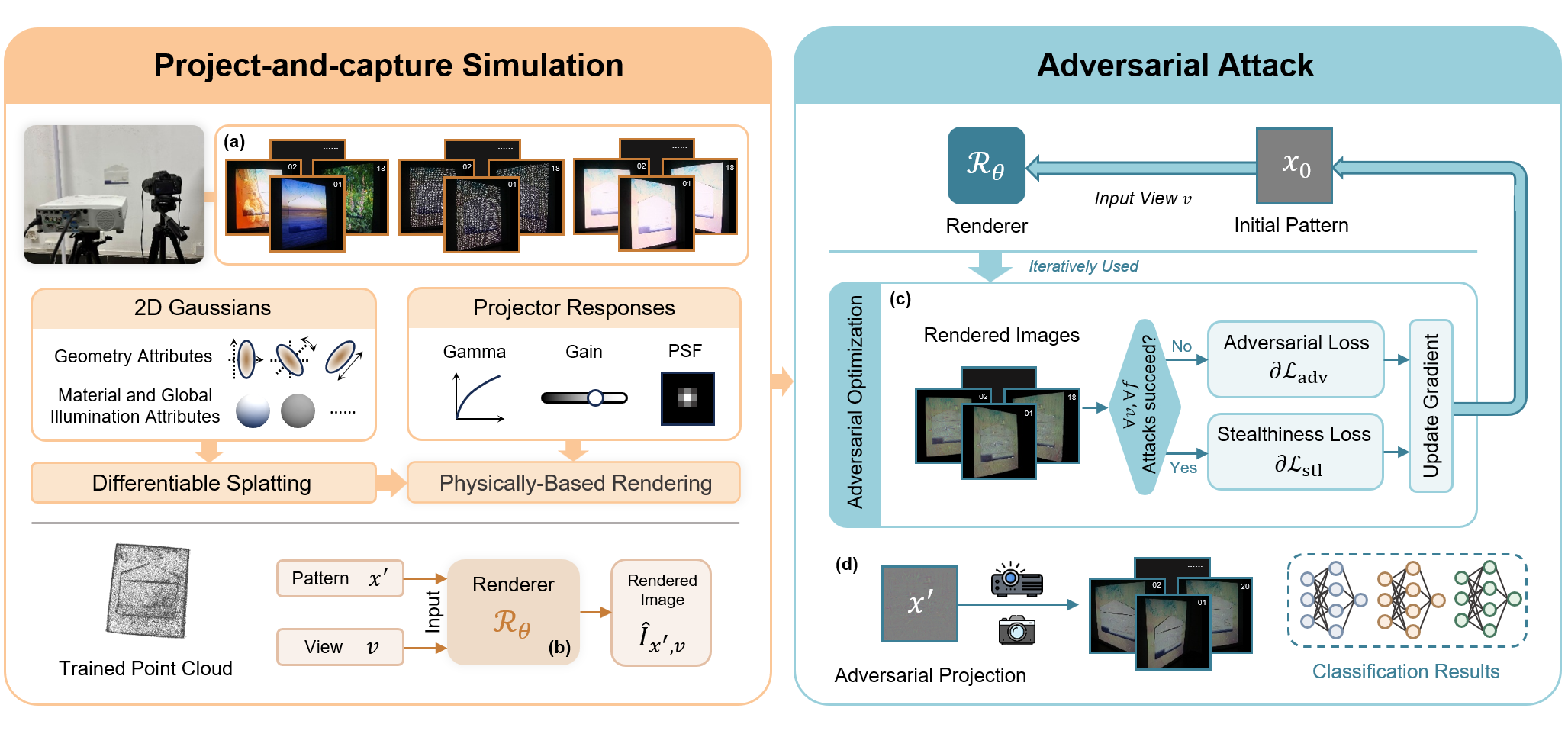}
\caption{
    The system overview of our \textbf{VIPA}. 
    The proposed framework is presented as two stages, \emph{Project-and-capture Simulation} and \emph{Adversarial Attack}, which are distinguished by different colors. 
    In the Project-and-capture Simulation stage, \textbf{(a)} we acquire training data by projecting, at each of 18 training viewpoints, a set of patterns and capturing the corresponding images. \textbf{(b)} These captures are then utilized to train a differentiable renderer $\mathcal{R}_\theta$ that jointly models the scene’s appearance and the projector–camera imaging process. Once trained, $\mathcal{R}\theta$ is used as a forward simulator that, given a camera viewpoint $v$ and a projector pattern $x'$, synthesizes the rendered image $\hat{I}_{x',v_i}$.
    In the Adversarial Attack stage, \textbf{(c)} starting from a plain gray initial pattern $x_0$, we iteratively refine the adversarial projection pattern through adversarial optimization. At each iteration, the rendered images $\{\hat{I}_{x',v_i}\}_{i=1}^{V}$ from all considered viewpoints $v_i$ are evaluated by multiple classifiers, and 
    if the minimum classifier confidence is below the target threshold or the perturbation remains within the stealthiness budget, a classifier-agnostic adversarial-loss gradient is applied to update $x'$; otherwise, it is updated using the stealthiness loss gradient, as specified in \autoref{alg: VIPA}. \textbf{(d)} Finally, the optimized adversarial projection $x'$ is physically deployed by projecting it onto the target scene, and its effectiveness is verified by randomly sampling 20 viewpoints to evaluate real-world transferability and the resulting classification outcomes.
    }
\label{fig:overall}
\end{center}
\end{figure*}

\subsection{Adversarial attack}
Adversarial attacks on deep image classifiers are typically categorized into digital and physical domains. Digital attacks perturb the input image to alter the prediction of a model, either steering it toward a chosen label (targeted) or away from the ground truth (untargeted). Most methods rely on back-propagating adversarial loss gradients to the input, ranging from efficient single-step approaches \cite{goodfellow2015explaining} to more powerful iterative schemes \cite{kurakin2017adversarial, madry2018towards, carlini2017}. While these gradient-based techniques assume white-box access to model parameters, gradient-free alternatives \cite{su2019one, zhao2019design} relax this requirement but suffer from reduced efficiency, particularly as input dimensionality increases.

Bringing adversarial examples into the physical world, by printing patches \cite{brown2018adversarialpatch}, affixing stickers \cite{li2019adversarial}, or modifying objects, introduces further complications because real capture processes can easily erase subtle perturbations. To mitigate this, robustness-oriented methods explicitly optimize over expected transformations: EOT \cite{athalye2018synthesizingrobustadversarialexamples} averages adversarial objectives across synthetic variations, $\text{RP}_2$ \cite{eykholt2018robust} tailors patches to real-world perturbations, and subsequent work \cite{jan2019connecting} models more complex digital-physical mappings. Yet, despite these advances, producing physically stealthy adversarial patterns remains a persistent challenge: robust examples often rely on conspicuous textures or shapes, while subtle perturbations rarely survive the distortions of real capture.

\noindent\textbf{Projector-based adversarial attack}. Unlike traditional physical-world attacks \cite{brown2018adversarialpatch, Sharif2016, athalye2018synthesizingrobustadversarialexamples, zhang2024} that alter objects or environments, projector-based adversarial attacks introduce perturbations by dynamically adjusting lighting conditions through projectors. By projecting a carefully designed adversarial projection, the scene captured by the camera can be misclassified by the classifier.

Early attempts include the work by Nichols and Jasper \cite{nichols2018projectingtroublelightbased}, which utilized a low-resolution 32×32 projector-camera setup to explore single-pixel adversarial perturbations. By employing a differential evolution (DE) algorithm \cite{storn1997differential} within the one-pixel attack framework \cite{su2019one}, they optimized only five parameters (pixel position and RGB values), demonstrating the feasibility of light-based adversarial perturbations.

Subsequent studies have aimed to improve stealthiness and practicality. Some works, particularly those targeting face recognition systems \cite{li2020lighthackfaceblackbox, nguyen2020Face, VLA2019, zhou2018invisiblemaskpracticalattacks}, employed custom hardware setups to enhance attack imperceptibility. Although effective in specialized settings, these methods often rely on narrow attack domains and special equipment, limiting their generalizability.

More recent methods, such as OPAD \cite{OPAD2021} and adversarial color projection \cite{hu2023adversarial}, emphasize practical attack deployment using projectors and rely on differentiable optimization pipelines. However, these methods often struggle to balance attack success rate with stealthiness, particularly in complex scenes or varying lighting conditions. To address this, SPAA \cite{SPAA2022} further advanced this line of research by explicitly modeling the project-and-capture pipeline with a differentiable neural renderer, significantly improving stealthiness and robustness. However, it remains constrained to single-classifier scenarios with fixed camera viewpoints, a significant limitation given the increasing adoption of multi-view and ensemble classification systems in real-world applications \cite{guo2020ensemble}. Li et al. \cite{CAPAA2025} introduces a novel classifier‑agnostic loss function and an attention-based gradient weighting mechanism that aggregates adversarial and stealthiness gradients across multiple classifiers. While significantly improving transferability and robustness across classifier architectures, it still falls short of achieving view-invariant attacks. We summarize the characteristics of different stealthy projector-based adversarial attack methods as \cref{tab:comparison}. We refer interested readers to \cite{hiPAA2024} for detailed review.

\subsection{ProCams simulation}
Simulation of projector-camera systems (ProCams) aims to model the physical process of projection and capture within a computational environment, enabling photorealistic image generation under novel projected patterns without requiring actual hardware. 

Early work in this domain primarily employed the Light Transport Matrix (LTM) \cite{LTM-ICRA2018, LTM-CVPR2014, LTM-SIGGRAPH2009, LTM-ACM2010}, which models the relationship between projector and camera pixels as a linear system. Although LTM can accurately capture the global illumination effect, its high computational cost, intensive sampling requirements and strict control of the environment limit its scalability and application.

To address these limitations, learning-based methods have been introduced. Huang et al. \cite{deprocams2021} proposed DeProCams, an end-to-end differentiable framework that learns both photometric and geometric mappings directly from captured data, thereby reducing reliance on explicit calibration. However, its formulation is restricted to fixed viewpoints and requires checkerboard-based calibration, limiting generalization. Similarly, Park et al. \cite{park2022projector} leverage differentiable rendering to model ProCams, but their approach depends on RGB-D input and per-pixel bias correction, which constrains applicability. Erel et al. \cite{erel2023neural} integrate projectors into Neural Radiance Fields (NeRFs) to achieve multi-view projection mapping. Despite enabling view-agnostic simulation, this approach assumes a camera-co-located light source and neglects ambient illumination, reducing its effectiveness in realistic environments.

A hybrid formulation is explored in DPCS \cite{dpcs2025}, which integrates geometric calibration, analytic reflectance modeling, and differentiable simulation into a unified optimization framework. By decoupling geometric alignment, photometric transfer, and material parameters, it offers a more interpretable alternative to purely neural rendering approaches. However, its reliance on parameterized scene representations and per-scene optimization limits its scalability to complex materials and novel viewpoints. Most recently, Deng et al. \cite{gsprocams2025} propose GS-ProCams, a Gaussian Splatting-based framework that reconstructs 3D scenes from multi-view project-and-capture data. This method supports efficient, differentiable rendering under novel projector inputs while explicitly modeling both geometric and photometric interactions in the scene. Compared to prior LTM and NeRF-based approaches, Gaussian Splatting achieves a favorable balance between computational efficiency, physical realism, and scene generalization.

\section{Methods}\label{sec:methods}

\subsection{Problem formulation}
Consider an image classifier $f$ that maps an input image $I$ to a predicted probability vector $ f(I) \in [0, 1]^N $, where $ f_i(I) $  represents the probability of the $i$-th class. Adversarial attacks aim to introduce a small perturbation $\delta$ to $I$, such that the perturbed image $I+\delta$ leads the classifier to misclassify:
\begin{align}\label{eq:adv_attack_formulation}
    &  \underset{i}{\arg\max} \, f_i(I + \delta)
    \begin{cases}
         = t, & \text{targeted attack} \\
         \neq t_{\text{true}}, & \text{untargeted attack}
    \end{cases} \nonumber \\
    & \quad\quad \text{subject to} \quad \mathcal{D}\left(I, I + \delta\right) < \epsilon, 
\end{align}
where $t_{\text{true}}$ is the true label for the object, $t$ is a target class for targeted attacks which is different from the true label, $\mathcal{D}$ is a similarity metric ensuring the perturbation remains imperceptible, and $\epsilon$ is a small threshold controlling the attack's stealthiness.

Extending \autoref{eq:adv_attack_formulation} to the physical world, projector-based attacks introduce perturbations by projecting a designed light pattern onto the physical scene. Denote the physical scene as $s$, the projector’s projection process as $\pi_p$, which maps an input image $x$ to a physical light field $\pi_p(x)$. This projected light interacts with the scene $s$, and the resulting image is captured by a camera with viewpoint $v$ through the capture process of camera $\pi_c$: 
\begin{align}\label{eq:sim_pc_formulation}
    &  I_{x,v} = \pi_c\left(\pi_p(x), s, v\right).
\end{align}

The goal of projector-based adversarial attacks is to find a perturbation $x'$ such that, when projected onto the scene and captured by a camera from pose $v$, the resulting image $I_{x',v}$ causes a classifier $f$ to output an incorrect prediction:
\begin{align}\label{eq:prj_adv_attack_formulation}
    &  \underset{i}{\arg\max} \, f_i(I_{x', v})
    \begin{cases}
         = t, & \text{targeted attack} \\
         \neq t_{\text{true}}, & \text{untargeted attack}
    \end{cases} \nonumber \\
    & \quad\quad \text{subject to} \quad \mathcal{D}\left(I_{x',v}, I_{x_0, v}\right) < \epsilon,
\end{align}
where $I_{x_0, v}$ is the camera-captured scene illuminated by a gray-light pattern $x_0$, without adversarial projection.

\subsection{View-invariant projector-based adversarial attack}
\noindent\textbf{Project-and-capture simulation.}
Inspired by previous studies \cite{gsprocams2025, hong2025gausstrap}, our method leverages a physically grounded projector-camera simulation framework based on Gaussian Splatting (GS-ProCams), enabling differentiable rendering of photorealistic images across diverse viewpoints.

The physical scene $s$ is represented as a collection of 2D Gaussian primitives. Each primitive is defined by its geometric attributes (location $p$, rotation $q$, scale $e$, and opacity $o$), and is further enriched with learnable material properties, including albedo $b$ and roughness $r$. Let the projector’s luminous power be $G_p$, when a projection pattern $x$ is projected, it undergoes the projector's radiometric and optical transformations. The direct incident radiance arriving at a surface point $x_s$ from the corresponding projector pixel coordinate $x_p$ is modeled as:
\begin{align}\label{eq:gs_Lp}
L_\text{in}(x_s) = \kappa \ast \left( G_p \cdot \left[ x(x_p) \right]^{\gamma_p} \right),
\end{align}
where $\kappa$ is the projector’s point-spread function, $\ast$ denotes convolution operation, and $\gamma_p$ denotes the projector's gamma decoding factor, responsible for transforming the input pattern into a linear color space.

\setlength{\algomargin}{0em}
\begin{algorithm}[t]
  \SetAlCapHSkip{0em}
  \SetAlgoLined 
  \SetKwComment{Comment}{$\triangleright$\ }{} 
  \SetSideCommentLeft
  \SetAlgoLongEnd
  \DontPrintSemicolon
  \SetKwInput{Input}{Input}
  \SetKwInOut{Output}{Output}
  \SetKw{KwOr}{or}
  \caption{VIPA: View-Invariant Projector-Based Adversarial Attack}\label{alg: VIPA}
  \Input{\;
  $x_0$: projector plain gray image\;
  $\mathcal{R}_{\theta}$: simulated project-and-capture process\;
  $\mathcal{F} = \{f_k\}_{k=1}^{K}$: a set of classifiers\;
  $\mathcal{V} = \{v_i\}_{i=1}^{V}$: a set of camera viewpoints\;
  $T$: number of optimization iterations\;
  $p_{\text{thr}}$: threshold for adversarial confidence\;
  $d_{\text{thr}}$: color perturbation threshold\;
  $\beta_1, \beta_2$: step sizes for adversarial and stealthiness losses\;
  }
  \Output{$x'$: projector adversarial image}
  Initialize $x' \gets x_0$\;
  \For{$t \gets 1$ \KwTo $T$}{
    
    $\hat{I}_{x',v_i} \gets \mathcal{R}_{\theta}(x', v_i),\quad \forall v_i \in \mathcal{V}$ \;
    $d \gets \frac{1}{|\mathcal{V}|} \sum_{v_i \in \mathcal{V}} \mathcal{D}\left(\hat{I}_{x',v_i}, I_{x_0, v_i}\right)$\;
    \eIf{$\min_{f_k \in \mathcal{F},\ v_i \in \mathcal{V}} f_k(x',v_i) < p_{\text{thr}}$ \KwOr $d<d_{\text{thr}}$}{
        $g_1 \gets \alpha\nabla_{x'} \mathcal{L}_{\text{adv}}$\ \Comment*[r]{min. adversarial loss}
        $x' \gets x' + \beta_1 \cdot \frac{g_1}{\|g_1\|_2}$ 
    }{
        $g_2 \gets -\nabla_{x'} \mathcal{L}_{\text{stl}}$\ \Comment*[r]{min. stealthiness loss}
        $x' \gets x' + \beta_2 \cdot \frac{g_2}{\|g_2\|_2}$ 
    }

    $x' \gets \text{clip}(x', 0, 1)$
  }

  \KwRet $x'$ that is adversarial and has smallest $d$\;
\end{algorithm}

Upon reaching the surface, the incoming illumination interacts with the object’s geometry and material. The outgoing radiance from $x_s$ toward the camera viewing direction $\omega_o$ is given by:
\begin{align}\label{eq:gs_Cp}
L_\text{out}(\omega_o, x_s) = f_{\mathrm{BRDF}} \cdot L_\text{in} \cdot (\omega_p \cdot n_s),
\end{align}
where $f_{\mathrm{BRDF}}$ is the bidirectional reflectance distribution function parameterized by the learned Gaussian attributes, $n_s$ is the surface normal, and $\omega_p$ denotes the incident light direction from the projector.

Finally, the camera-captured pixel intensity combines the direct reflected radiance $L_{\text{out}}$ with an approximated global illumination component, $L_{\text{amb}}$. To efficiently account for environmental and indirect lighting, $L_{\text{amb}}$ is modeled as a view-dependent residual color parameterized by spherical harmonics, which is accumulated along the camera ray via differentiable alpha-blending rasterization. By merging these components and applying the camera’s tone mapping $\gamma_c$ and color clamping, the intensity is given by:
\begin{align}\label{eq:gs_I}
    \hat{I}_{x, v}=\operatorname{clamp}\left((L_\text{out} + L_\text{amb})^{\frac{1}{\gamma_c}},0,1\right),
\end{align}
where $\gamma_c$ is the camera’s gamma correction parameter, and $v \in \mathcal{V}$ is the camera viewpoint. Denote the project-and-capture simulation model in \autoref{eq:gs_I} as $\mathcal{R}_{\theta}$, we have 
\begin{align}\label{eq:gs_R}
    \hat{I}_{x,v} = \mathcal{R}_{\theta}(x, v).
\end{align}

To train the parameters $\theta$ of GS-ProCams $\mathcal{R}_{\theta}$, we begin by capturing the target scene from a set of $V$ camera poses $\{v_i\}_{i=1}^{V}$ , each under multiple projector inputs: black, calibration, white, gray, and a color pattern that is different for different views. We empirically set the number of training viewpoints to 18, balancing reconstruction quality and acquisition reliability, as excessive views may cause COLMAP matching failures due to projection distortions. This allows the model to disentangle the scene's intrinsic properties (geometry and materials) from the projector's and environment's lighting. The optimizable parameters, including the Gaussian attributes $\left \{ p,q,e,o,b,r \right \}$ and the projector responses $ \left\{\gamma_{p}, G_{p}, \kappa\right\}$, are jointly optimized over 20,000 iterations, yielding a physically accurate and differentiable project-and-capture simulation that is later used for adversarial optimization.

\vspace{.5em}
\noindent\textbf{Adversarial attack. }
Once GS-ProCams $\mathcal{R}_{\theta}$ is fully trained, we can perform simulated adversarial attack. The objective is to generate a single universal adversarial projection pattern $x'$ such that, when it is projected onto the scene, the resulting captured images from various viewpoints lead to misclassification by a diverse set of classifiers.

Let $\mathcal{F} = \{f_k\}_{k=1}^{K}$ denote a set of pre-trained deep image classifiers, and $\mathcal{V} = \{v_i\}_{i=1}^{V}$ denote a set of camera viewpoints. The projection pattern $x'$ is projected onto the scene, and under each viewpoint $v_i$, a corresponding image $\hat{I}_{x', v_i}$ is rendered by the trained GS-ProCams model. The adversarial objective is defined as follows:
\begin{align}\label{eq:attack_express} 
    & \forall f_k \in \mathcal{F},\ \forall v_i \in \mathcal{V}, \nonumber\\ 
    & {\arg\max}\, f_k(I_{x', v_i}) 
    \begin{cases}
         = t, & \text{targeted attack} \\
         \neq t_{\text{true}}, & \text{untargeted attack}
    \end{cases}  \nonumber\\
    & \quad\quad\text{subject to} \quad \mathcal{D}\left(I_{x', v_i}, I_{x_0, v_i}\right) < \epsilon.
\end{align}
Here, we choose $L_2$ norm as the image distance/stealthiness metric $\mathcal{D}$.

As the captured scene may be interpreted differently by various classifiers and views, especially in cases where visually similar categories exist. For instance, a clock can be recognized as a “wall clock” or an “analog clock” depending on the bias of the classifier or the specific viewing angle. To establish the ground-truth label $t_{\text{true}}$ for each classifier in a robust, view-consistent manner, we apply a majority vote strategy across all views. Specifically, given a set of original images $\{{I_{x_0, v_i}}\}_{i=1}^{V}$ captured under a plain gray projection pattern $x_0$, we define $t_{\text{true}}^{(k)}$ for the classifier $f_k$ as the original class among all views:
\begin{align}\label{eq:org_label} 
    t_{\text {true }}^{(k)}=\underset{m \in\{1, \ldots, M\}}{\operatorname{argmax}} \sum_{i=1}^{V} \mathcal{I}\left[f_{k}\left(I_{x_{0}, v_{i}}\right)=m\right],
\end{align}
where $\mathcal{I}[\cdot]$ is an indicator function that returns 1 if the condition holds and 0 otherwise, $m$ denotes a candidate class label in the classifier’s label set, and $M$ is the total possible classes for classifier $f_k$.

To balance attack effectiveness and imperceptibility, we jointly minimize the adversarial loss and stealthiness loss across all views:
\begin{align}\label{eq:adv_prj_exp}
x' = \arg\min_{x'} \;  \alpha\mathcal{L}_{\text{adv}}\left(\hat{I}_{x', v_i}\right) + \mathcal{D}\left(\hat{I}_{x', v_i}, I_{x_0, v_i}\right),
\end{align}
where $v_i \in \mathcal{V}$, $\alpha=-1$ for targeted attacks and $\alpha=1$ for untargeted attacks. 

The adversarial loss, $\mathcal{L}_{\text{adv}}$, encourages the projected pattern $x'$ to drive classifier predictions toward a target class for targeted attack or away from the original label in untargeted attack. We calculate it over a set of classifiers $\mathcal{F}=\{f_1,f_2,...,f_k\}$ and a set of camera viewpoints $\mathcal{V}=\{v_1,v_2,...,v_k\}$. Denote {$z^{(k)}_{t_\text{true}^{(k)}}(\cdot)$} as the $k$-th classifier's output logit (raw classification score) corresponding to the true class $t_\text{true}^{(k)}$. For untargeted attacks, our objective is to explicitly minimize this true-class logit. Thus, the adversarial loss is computed as:
\begin{align}\label{eq:untar_adv_loss}
 	&   \mathcal{L}_{\text{adv}} = \frac{1}{|\mathcal{V}|} \sum_{v_i \in \mathcal{V}} \sum\nolimits_{k = 1}^{K} \omega_{k} \cdot z^{(k)}_{t_\text{true}^{(k)}}\left(\hat{I}_{x', v_i}\right),
\end{align}
where $\omega_k$ is the assigned weight of classifier $f_k$, which can be tuned to reflect its relative importance~\cite{CAPAA2025}.

As for targeted attack, $\mathcal{L}_{\text{adv}}$ is expressed as:

\begin{equation}\label{eq:tar_adv_loss}
\resizebox{0.87\linewidth}{!}{$
    \displaystyle
        \mathcal{L}_{\text{adv}} \;=\;
        \frac{1}{|\mathcal{V}|}
        \sum_{v_i \in \mathcal{V}}
        \sum_{k=1}^{K}
        \omega_{k}\,
        \log\!\left(\operatorname{softmax}\!\bigg(\frac{z^{(k)}_{t}\!\big(\hat{I}_{x',v_i}\big)}{T}\bigg)\right),
$}
\end{equation}
where the parameter $T$ acts as a temperature parameter and is dynamically adjusted over optimization steps.

While maximizing the adversarial objective, we should simultaneously ensure that the adversarial pattern remains imperceptible or visually plausible when projected. We compute the stealthiness loss through ${\mathcal{L}_\mathrm{prjL2}}$, $\mathcal{L}_{\mathrm{camL} 2}$ and $\mathcal{L}_{\mathrm{camdE}}$ across all views as \autoref{eq:stl_loss}:

\begin{equation}\label{eq:stl_loss}
\resizebox{0.87\linewidth}{!}{$
\displaystyle
    \mathcal{L}_{\text{stl}}
    = w_{1}\cdot \mathcal{L}_{\mathrm{prjL2}}
    + \frac{w_{2}}{|\mathcal{V}|}\sum_{v_i \in \mathcal{V}} \mathcal{L}_{\mathrm{camL2}}^{(v_i)}
    + \frac{w_{3}}{|\mathcal{V}|}\sum_{v_i \in \mathcal{V}} \mathcal{L}_{\mathrm{camdE}}^{(v_i)},
    $}
\end{equation}
where $w_1$, $w_2$, $w_3$ are the weights for ${\mathcal{L}_\mathrm{prjL2}}$, $\mathcal{L}_{\mathrm{camL} 2}$ and $\mathcal{L}_{\mathrm{camdE}}$, respectively.

The optimization is carried out for $T = 30$ iterations based on empirical observation, which achieves stable convergence across all settings, as smaller values may lead to occasional non-convergence while larger values bring no noticeable improvement. We initialize $x'$ as a plain gray pattern and update it using two learning rates: $\beta_1 = 3$ for adversarial loss and $\beta_2 = 1$ for stealthiness loss. And, at each iteration, we render all 18 camera viewpoints $\{I_{x',v_i}\}_{i=1}^{V}$ under the current $x'$, as \autoref{alg: VIPA} shows.

\section{Experiments}\label{sec:experiments}

\subsection{Experimental setup}
\noindent\textbf{System Configuration.} 
Our setup consists of a Canon EOS 600D DSLR camera and an EPSON CB-965 3LCD projector, as shown in \cref{fig:teaser}~(a), and their resolutions are both set to 800 × 800. Considering projector-camera synchronization, we manually set a 200-ms delay between the time a pattern is projected and the moment the camera captures the image. We deploy VIPA on a laptop with Intel Core i7-13700H CPU, 16GB memory, and an Nvidia GeForce RTX 4060 Laptop GPU.

\vspace{.2em}
\noindent\textbf{Datasets and Implementation.}
We captured five distinct real-world setups featuring objects ranging from flat surfaces to three-dimensional shapes to ensure scene diversity. All quantitative comparisons and qualitative evaluations in the main experiments are conducted on these five setups. The VIPA framework is trained individually for each scene, taking approximately 15 minutes for training and about 30 to 45 minutes for the attacking process, where the relatively high cost is mainly due to 30 optimization iterations, 18 rendered viewpoints per iteration, and joint optimization over three classifier losses. We feed the camera-captured adversarial scenes into three classifiers, Inception v3 \cite{2016inceptionv3}, ViT \cite{2021ViT} and Convnext v2 \cite{woo2023convnextv2}, for real-world projector-based adversarial attack evaluation. In addition, we conduct a view ablation study to investigate the impact of the number of attack viewpoints during optimization. Since the original physical setups used for the main experiments were dismantled after data collection and could not be fully reconstructed, the view ablation experiments were performed on three newly constructed real-world setups instead. These setups are different from the five setups used in the main experimental comparisons, while following the same capture and evaluation protocol. We vary the number of attack views as 1, 5, 9, 12, and 18 during optimization and evaluate their effects on both attack performance and stealthiness under untargeted and targeted attack settings.

\vspace{.2em}
\noindent\textbf{Baselines.} 
We compare our method with three baselines: (1) SPAA \cite{SPAA2022}, a view-specific and classifier-specific approach; (2) CAPAA \cite{CAPAA2025}, which is classifier-agnostic but not view-invariant; and (3) VIPA classifier-specific, a variant of our method tailored to specific models. Since SPAA \cite{SPAA2022} and VIPA classifier-specific cannot perform classifier-agnostic attack, we compare them in a classifier-specific manner. We allow them attack each classifier individually, which means they will generate three different adversarial projections according to the three different classifiers we use instead of only generating one pattern to fool all the classifiers.

\begin{table}[tb]
\centering
\caption{
Detailed camera parameters of untargeted attack test views on envelope. For each view, we report the camera–scene distance together with the horizontal and vertical angles, and indicate their corresponding directions. The visualization is shown in \cref{fig:viewspace}.
}
\label{tab:camera_para_envelope}
\begin{tabular}{@{}r@{\hskip 6pt}c@{\hskip 6pt}rc@{\hskip 6pt}rc@{}}
    \toprule[0.5mm]
    \multirow{2}{*}{\textbf{View}} & \multirow{2}{*}{\textbf{Distance (m)}} & \multicolumn{2}{c}{\textbf{Horizontal}} & \multicolumn{2}{c}{\textbf{Vertical}} \\ 
    \cmidrule(lr){3-4} \cmidrule(lr){5-6}
     &  & \textbf{Angle} & \textbf{Direction} & \textbf{Angle} & \textbf{Direction} \\ 
    \midrule
    1  & 0.652 & 10° & Right & 3° & Up \\
    2  & 0.707 & 18° & Right & 12° & Up \\
    3  & 0.727 & 38° & Right & 2° & Up \\
    4  & 0.657 & 31° & Right & 2° & Up \\
    5  & 0.640 & 26° & Right & 7° & Up \\
    6  & 0.721 & 47° & Left & 16° & Up \\
    7  & 0.723 & 34° & Right & 2° & Up \\
    8  & 0.693 & 38° & Left & 1° & Down \\
    9  & 0.677 & 37° & Right & 6° & Up \\
    10 & 0.730 & 47° & Right & 10° & Up \\
    
    11 & 0.731 & 41° & Left & 16° & Up \\
    12 & 0.679 & 21° & Right & 18° & Up \\
    13 & 0.678 & 42° & Left & 2° & Down \\
    14 & 0.750 & 19° & Left & 3° & Up \\
    15 & 0.732 & 55° & Left & 14° & Up \\
    16 & 0.634 & 21° & Right & 7° & Up \\
    17 & 0.644 & 24° & Right & 2° & Up \\
    18 & 0.704 & 57° & Left & 9° & Up \\
    19 & 0.643 & 15° & Right & 12° & Up \\
    20 & 0.701 & 31° & Left & 14° & Up \\
    \bottomrule[0.5mm]
\end{tabular}
\end{table}
\begin{figure}
\centering
\includegraphics[width=1.0\linewidth]{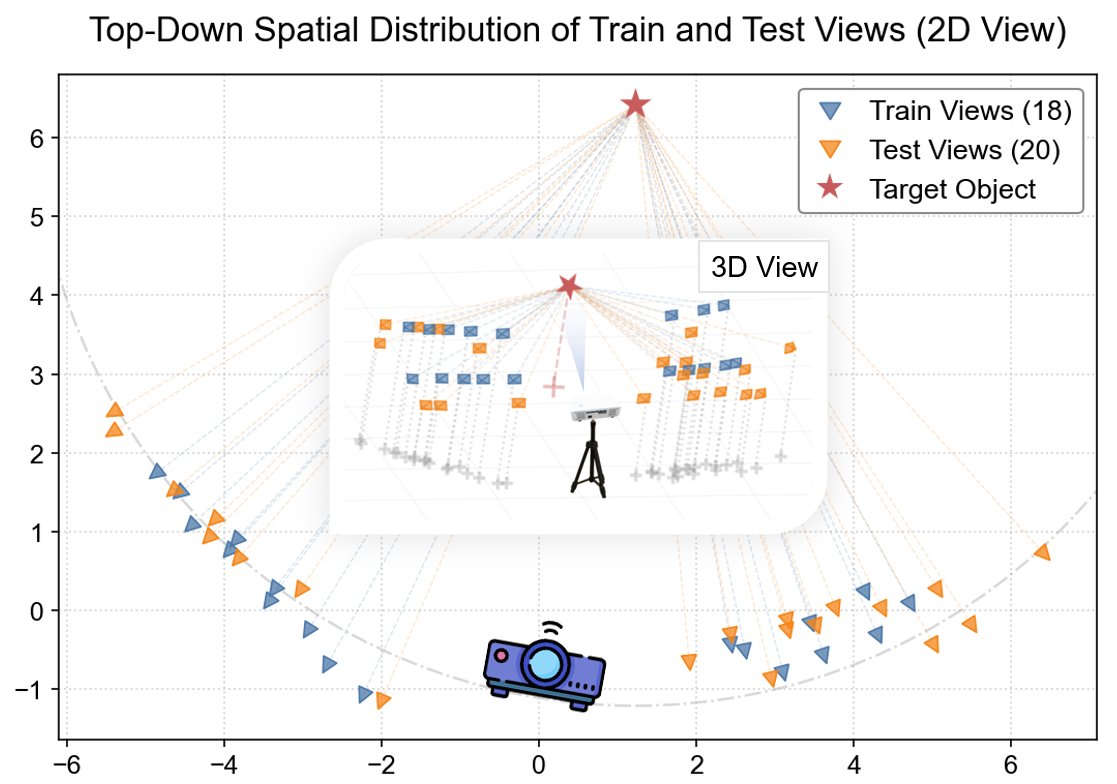}
\caption{The visualization of 18 training views and 20 test views for untargeted attack on the envelope scene. The main plot illustrates the top-down horizontal distribution of the camera positions relative to the target scene, where blue circles and orange triangles denote training and test viewpoints, respectively. The red pentagram represents the entire scene, scaled down to a single point for geometric reference. The 3D View inset provides a spatial visualization of the corresponding capture configuration, where camera frustums indicate viewing directions and the gray cross points mark the vertical projection of each camera's viewpoint onto the horizontal plane.}
\label{fig:viewspace}
\end{figure}

\begin{figure*}[!t]
	\begin{center}
		\includegraphics[width=1\linewidth]{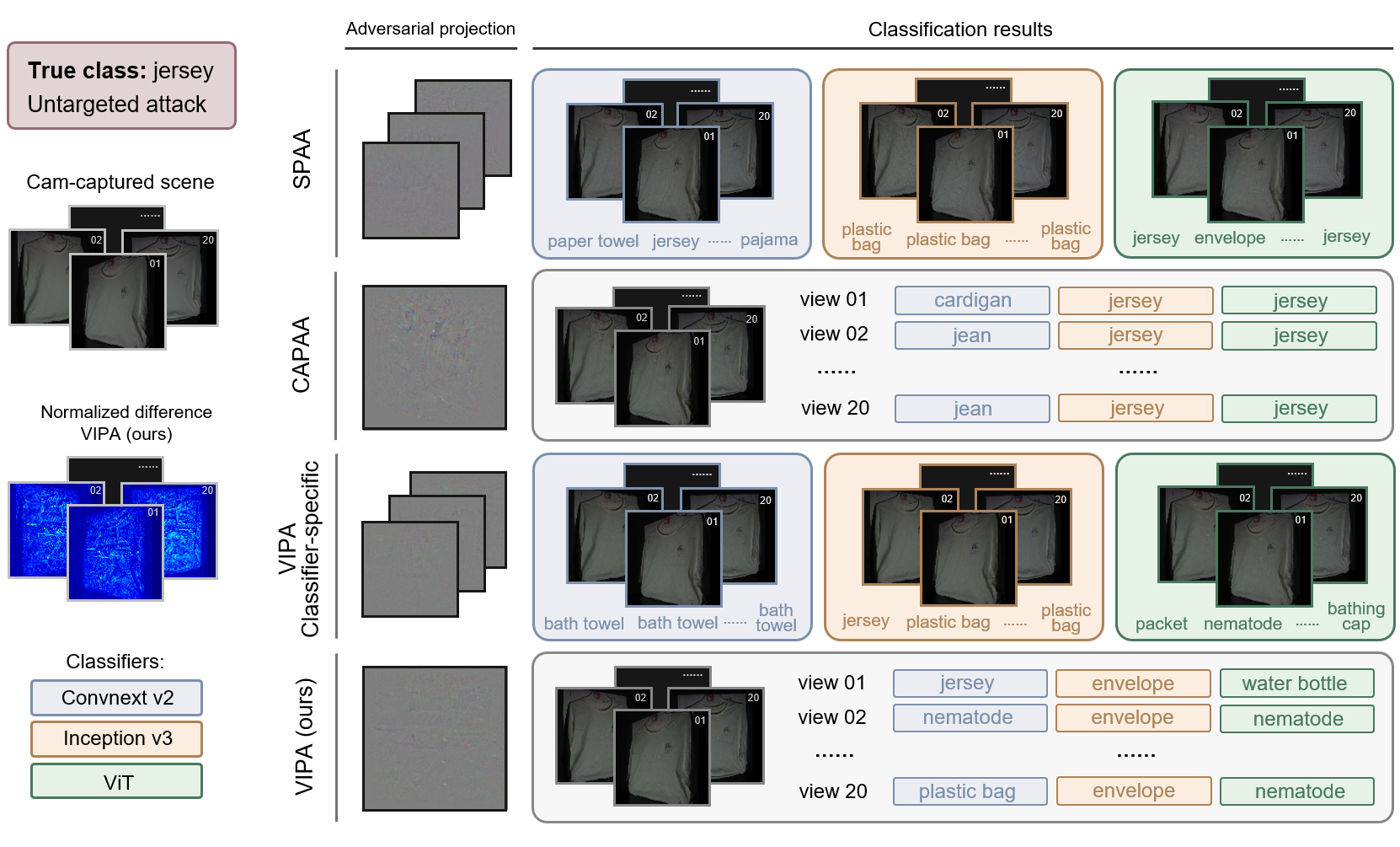}
		\caption{
        \textbf{Untargeted} projector-based adversarial attack. The goal is to use adversarial projections to induce various classifiers to misclassify the camera-captured scene as any label other than the true class, \textbf{jersey}, from various views. For evaluation, we randomly sample and capture 20 test viewpoints. On the far left, we present the clean camera-captured scene and the corresponding normalized difference maps computed against the camera-captured images obtained after projecting the adversarial pattern produced by our method. For each approach, the 1\textsuperscript{st} column visualizes its adversarial projections. Note that CAPAA \cite{CAPAA2025} and VIPA (ours) are classifier-agnostic and therefore produce a single adversarial pattern for all classifiers, whereas VIPA classifier-specific and SPAA \cite{SPAA2022} generate distinct adversarial patterns for each of the three classifiers. In the classification results, we consider three classifiers, Convnext v2 \cite{woo2023convnextv2}, Inception v3 \cite{2016inceptionv3} and ViT \cite{2021ViT}, which are denoted in \protect\colorboxblock{cblue_border}{cblue_fill},  \protect\colorboxblock{corange_border}{corange_fill} and \protect\colorboxblock{cgreen_border}{cgreen_fill}, respectively, and annotate each classifier’s predicted label next to the corresponding viewpoint.
        }\label{fig:jersey_untar}
	\end{center}
\end{figure*}

\begin{table*}[!t]
\vspace{-2mm}
\centering
\fontsize{7.6pt}{11pt}\selectfont
\caption{
Quantitative comparison of different attack methods under both untargeted and targeted attack scenarios. The 1$^\text{st}$ and 2$^\text{nd}$ columns report the attack method and the threshold parameter (d$_{\text{thr}}$), respectively. For each method, three consecutive rows correspond to (d$_{\text{thr}}$=4,5,6), and the Avg. OSR column gives the mean success rate across these three thresholds. Columns 3 to 8 summarize untargeted attack results, where columns 3 to 6 are stealthiness metrics of the camera-captured adversarial projections, and columns 7 to 8 report the attack success rate (OSR) and its average across thresholds. Columns 9 to 14 summarize the corresponding metrics under targeted attacks, which are the same as those for untargeted attacks. All results are aggregated over five experimental setups.
}\label{tab:table_compare}
\begin{tabular}{@{}lccccccccccccc@{}}
\toprule[0.5mm]
\multirow{2}{*}{\textbf{Method}} & \multirow{2}{*}{\textbf{d$_\textbf{thr}$}} & \multicolumn{6}{c}{\textbf{Untargeted Attack}} & \multicolumn{6}{c}{\textbf{Targeted Attack}} \\ \cmidrule(lr){3-8} \cmidrule(lr){9-14} 
 &  & \textbf{$\Delta \textbf{E}\downarrow$} & \textbf{L$_\infty\downarrow$} & \textbf{PSNR$\uparrow$} & \textbf{SSIM$\uparrow$} & $\textbf{OSR$\uparrow$}$ & $\textbf{Avg. OSR$\uparrow$}$ & \textbf{$\Delta \textbf{E}\downarrow$} & \textbf{L$_\infty\downarrow$} & \textbf{PSNR$\uparrow$} & \textbf{SSIM$\uparrow$} & $\textbf{OSR$\uparrow$}$ & $\textbf{Avg. OSR$\uparrow$}$ \\ \midrule[0.3mm]
\multirow{3}{*}{SPAA \cite{SPAA2022}} & 4 & 1.58 & 2.61 & 39.74 & 0.9687 & 57.67\% &  & 1.58 & 2.65 & 39.47 & 0.9679 & 8.00\% &  \\
 & 5 & 1.70 & 2.91 & 38.54 & 0.9625 & 54.67\% & 58.33\% & 1.76 & 3.01 & 38.14 & 0.9611 & 5.34\% & 7.22\% \\
 & 6 & 1.94 & 3.28 & 37.50 & 0.9557 & 62.67\% &  & 2.01 & 3.39 & 37.13 & 0.9551 & 8.33\% &  \\ \cmidrule[0.3mm]{1-14}
\multirow{3}{*}{CAPAA \cite{CAPAA2025}} & 4 & 1.64 & 2.76 & 39.21 & 0.9687 & 53.33\% &  & 1.67 & 2.79 & 39.26 & 0.9687 & 21.67\% &  \\
 & 5 & 1.64 & 2.76 & 39.20 & 0.9682 & 58.00\% & 56.89\% & 1.64 & 2.76 & 39.39 & 0.9694 & 17.67\% & 18.78\% \\
 & 6 & 1.64 & 2.77 & 39.13 & 0.9683 & 59.33\% &  & 1.66 & 2.77 & 39.31 & 0.9689 & 17.00\% &  \\ \cmidrule[0.3mm]{1-14}
\multirow{3}{*}{\shortstack[l]{VIPA \\classifier-specific}}  & 4 & 1.53 & 2.57 & 40.31 & 0.9745 & 67.00\% &  & 1.53 & 2.53 & 40.60 & 0.9756 & 24.00\% &  \\
 & 5 & 1.54 & 2.57 & 40.30 & 0.9745 & 65.33\% & 67.11\% & 1.53 & 2.54 & 40.57 & 0.9757 & 22.33\% & 21.78\% \\
 & 6 & 1.53 & 2.57 & 40.32 & 0.9744 & 69.00\% &  & 1.53 & 2.54 & 40.56 & 0.9757 & 19.00\% &  \\ \cmidrule[0.3mm]{1-14}
\multirow{3}{*}{VIPA (ours)} & 4 & 1.58 & 2.65 & 39.97 & 0.9727 & 71.33\% &  & 1.57 & 2.60 & 40.28 & 0.9740 & 32.33\% &  \\
 & 5 & 1.58 & 2.66 & 39.93 & 0.9727 & 76.33\% & \textbf{74.89\%} & 1.56 & 2.60 & 40.26 & 0.9740 & 29.00\% & \textbf{32.00\%} \\
 & 6 & 1.59 & 2.66 & 39.95 & 0.9729 & 77.00\% &  & 1.56 & 2.57 & 40.37 & 0.9740 & 34.67\% &  \\ \bottomrule[0.5mm]
\end{tabular}
\end{table*}

\subsection{Evaluation protocol}
To comprehensively assess the robustness and view-invariant property of our VIPA, we design a benchmarking protocol based on unseen viewpoints and standardized metrics.

\subsubsection{Test views sampling} 
We perform testing under viewpoints that are distinct from those used during the data capturing and training stages. These test views are randomly sampled to cover a diverse range of spatial configurations, including horizontal variations from left to right and vertical displacements ranging from upward-looking to downward-looking perspectives.

To facilitate intuitive understanding and spatial analysis of these test viewpoints, we leverage COLMAP \cite{COLMAP2016}, a widely adopted structure-from-motion and multi-view stereo pipeline, to reconstruct the 3D scene and visualize the test camera poses. Additionally, we quantitatively characterize each camera viewpoint using the following three geometric parameters:
\begin{itemize}
    \item Distance: The Euclidean distance between the camera center and the scene, measured in decimeters.
    \item Horizontal Angle: The azimuthal angle between the camera and the scene, labeled with left or right orientation.
    \item Vertical Angle: The elevation angle indicating whether the camera is tilted upward or downward with respect to the horizontal plane.
\end{itemize}

An example is presented in \cref{tab:camera_para_envelope}, showing the 20 test views' camera parameters of untargeted attack on the envelope scene. The corresponding visualization of it is shown in \cref{fig:viewspace}. Note that since our attack relies on a physical projector to project adversarial patterns onto the target, the sampling range for these test views is constrained to a 180-degree frontal hemisphere relative to the scene.

In comparative experiments, for one scene and at each test viewpoint, we sequentially project and capture adversarial patterns generated by different attack methods. This ensures that all methods are compared fairly at each randomly sampled viewpoint. Some visualizations of test viewpoints for scenes are provided in the supplementary materials.

\begin{figure*}[!t]
	\begin{center}
		\includegraphics[width=1\linewidth]{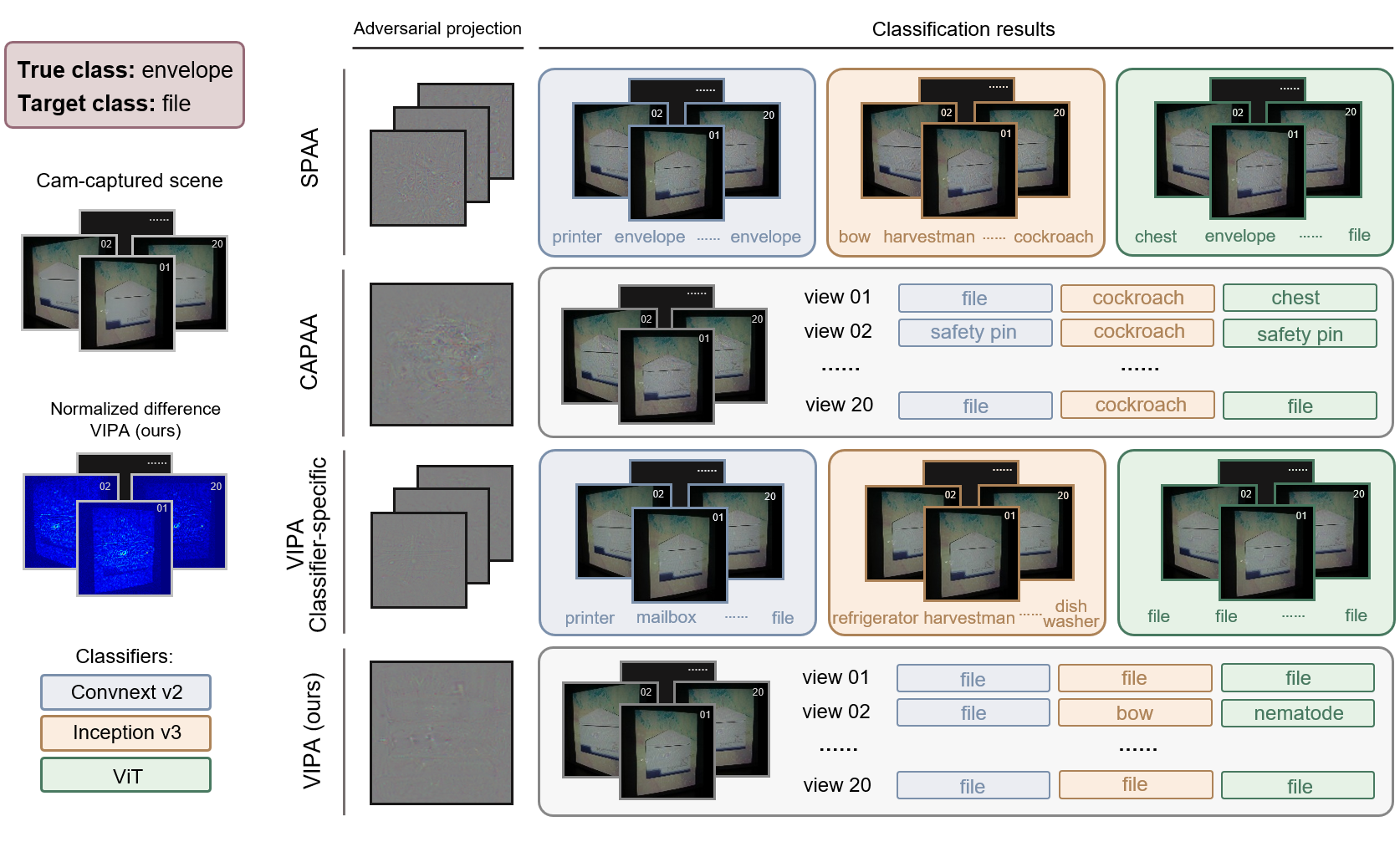}
		\caption{
        \textbf{Targeted} projector-based adversarial attack. The goal is to use adversarial projections to cause different classifiers to misclassify the camera-captured scene \textbf{envelope} as \textbf{file} from various viewpoints. For evaluation, we randomly sample and capture 20 test viewpoints. On the far left, we present the clean camera-captured scene and the corresponding normalized difference maps computed against the camera-captured images obtained after projecting the adversarial pattern produced by our method. For each approach, the 1\textsuperscript{st} column visualizes its adversarial projections. Note that CAPAA \cite{CAPAA2025} and VIPA (ours) are classifier-agnostic and therefore produce a single adversarial pattern for all classifiers, whereas VIPA classifier-specific and SPAA \cite{SPAA2022} generate distinct adversarial patterns for each of the three classifiers. In the classification results, we consider three classifiers, Convnext v2 \cite{woo2023convnextv2}, Inception v3 \cite{2016inceptionv3} and ViT \cite{2021ViT}, which are denoted in \protect\colorboxblock{cblue_border}{cblue_fill},  \protect\colorboxblock{corange_border}{corange_fill} and \protect\colorboxblock{cgreen_border}{cgreen_fill}, respectively, and annotate each classifier’s predicted label next to the corresponding viewpoint.
        }\label{fig:envelope_tar}
	\end{center}
\end{figure*}

\begin{table*}[!t]
\vspace{-3mm}
\centering
\fontsize{7.8pt}{11pt}\selectfont
\caption{
View ablation study on the effect of the number of attack views used during optimization under both untargeted and targeted projector-based attack settings. We vary the number of attack views from 1, 5, 9, 12 to 18 while keeping all other settings unchanged to evaluate the impact of viewpoint diversity on attack performance and stealthiness. The table reports results for different threshold values d$_{\text{thr}}$ = 4, 5, 6. Columns 3 to 6 present stealthiness metrics, and columns 7 to 8 report OSR and Avg. OSR for untargeted attacks. Columns 9 to 14 present the corresponding metrics for targeted attacks. All results are averaged over three experimental setups.
}\label{tab:views_ablation}
\vspace{-2.5mm}
\begin{tabular}{@{}cccccccccccccc@{}}
\toprule[0.5mm]
\multirow{2}{*}{\textbf{Attack Views}} & \multirow{2}{*}{\textbf{d$_\textbf{thr}$}} & \multicolumn{6}{c}{\textbf{Untargeted Attack}} & \multicolumn{6}{c}{\textbf{Targeted Attack}} \\ \cmidrule(lr){3-8} \cmidrule(lr){9-14} 
 &  & \textbf{$\Delta \textbf{E}\downarrow$} & \textbf{L$_\infty\downarrow$} & \textbf{PSNR$\uparrow$} & \textbf{SSIM$\uparrow$} & $\textbf{OSR$\uparrow$}$ & $\textbf{Avg. OSR$\uparrow$}$ & \textbf{$\Delta \textbf{E}\downarrow$} & \textbf{L$_\infty\downarrow$} & \textbf{PSNR$\uparrow$} & \textbf{SSIM$\uparrow$} & $\textbf{OSR$\uparrow$}$ & $\textbf{Avg. OSR$\uparrow$}$ \\ \midrule[0.3mm]
\multirow{3}{*}{1} & 4 & 1.45 & 2.30 & 41.48 & 0.9692 & 48.33\% &  & 1.39 & 2.19 & 42.09 & 0.9716 & 7.78\% &  \\
 & 5 & 1.45 & 2.32 & 41.38 & 0.9690 & 49.44\% & 50.74\% & 1.39 & 2.19 & 42.09 & 0.9717 & 10.56\% & 10.18\% \\
 & 6 & 1.45 & 2.32 & 41.39 & 0.9691 & 54.44\% &  & 1.39 & 2.19 & 42.09 & 0.9716 & 12.22\% &  \\ \cmidrule[0.3mm]{1-14}
\multirow{3}{*}{5} & 4 & 1.53 & 2.47 & 40.64 & 0.9653 & 68.89\% &  & 1.51 & 2.41 & 41.03 & 0.9674 & 22.78\% &  \\
 & 5 & 1.53 & 2.47 & 40.63 & 0.9651 & 66.67\% & 68.52\% & 1.51 & 2.41 & 41.03 & 0.9674 & 19.44\% & 20.74\% \\
 & 6 & 1.54 & 2.48 & 40.57 & 0.9649 & 70.00\% &  & 1.51 & 2.41 & 41.04 & 0.9675 & 20.00\% &  \\ \cmidrule[0.3mm]{1-14}
\multirow{3}{*}{9} & 4 & 1.56 & 2.53 & 40.31 & 0.9642 & 80.00\% &  & 1.54 & 2.47 & 40.65 & 0.9661 & 35.00\% &  \\
 & 5 & 1.56 & 2.52 & 40.33 & 0.9639 & 75.55\% & 79.26\% & 1.54 & 2.48 & 40.62 & 0.9659 & 27.22\% & 29.44\% \\
 & 6 & 1.56 & 2.52 & 40.32 & 0.9641 & 82.22\% &  & 1.54 & 2.47 & 40.65 & 0.9659 & 26.11\% &  \\ \cmidrule[0.3mm]{1-14}
\multirow{3}{*}{12} & 4 & 1.59 & 2.57 & 40.07 & 0.9635 & 82.22\% &  & 1.56 & 2.52 & 40.37 & 0.9654 & 36.11\% &  \\
 & 5 & 1.59 & 2.58 & 40.02 & 0.9632 & 80.00\% & 82.78\% & 1.56 & 2.52 & 40.31 & 0.9651 & 37.78\% & 38.89\% \\
 & 6 & 1.59 & 2.58 & 40.05 & 0.9635 & 86.11\% &  & 1.56 & 2.52 & 40.34 & 0.9651 & 42.78\% &  \\ \cmidrule[0.3mm]{1-14}
\multirow{3}{*}{18} & 4 & 1.62 & 2.64 & 39.71 & 0.9629 & 89.44\% &  & 1.59 & 2.57 & 40.00 & 0.9647 & 60.00\% &  \\
 & 5 & 1.61 & 2.63 & 39.74 & 0.9629 & 93.33\% & 90.93\% & 1.59 & 2.58 & 39.98 & 0.9644 & 47.22\% & 51.48\% \\
 & 6 & 1.62 & 2.64 & 39.70 & 0.9627 & 90.00\% &  & 1.59 & 2.57 & 40.03 & 0.9645 & 47.22\% &  \\ \bottomrule[0.5mm]
\end{tabular}
\end{table*}

\subsubsection{Evaluation metrics} 
We evaluate attack success rate using the Overall Success Rate (OSR), which is defined as \autoref{eq:OSR_define}:
\begin{equation}\label{eq:OSR_define}
\resizebox{0.87\linewidth}{!}{$
\displaystyle
    \text{OSR}=\frac{1}{|\mathcal{V}||\mathcal{F}|N}
    \sum_{v_i\in\mathcal{V}}\sum_{f_k\in\mathcal{F}}
    \mathcal{I}\!\left[
    f_k(I_{x',v_i})
    \begin{cases}
        =\, t, & \text{targeted}\\[2pt]
        \neq\, t_{\text{true}}, & \text{untargeted}
    \end{cases}
\right],
$}
\end{equation}
where $t$ denotes the targeted class we set for the scene in targeted attack, $t_{\text{true}}$ is the original true class of the scene, $\mathcal{V}$ and $\mathcal{F}$ denote the sets of test viewpoints and image classifiers respectively, $N$ is the number of different color perturbation threshold $d_\text{thr}$, $\mathcal{I}[\cdot]$ is the indicator function that returns 1 if the condition holds and 0 otherwise. We further report Avg. OSR, computed as the mean OSR over all tested $d_{\text{thr}}$ values, thereby reflecting the overall reliability of attacks.

To evaluate the stealthiness of the adversarial projection, we follow prior projector-based attack studies \cite{CAPAA2025, SPAA2022} and adopt four complementary stealthiness metrics, each representing a different aspect of perceptual deviation between the original captured scene $I_{x,v}$ and the adversarially captured image. 
Specifically, $\Delta$E measures perceptual color differences in the CIEDE2000 \cite{CIEDE2000} color space, reflecting how noticeable the projection-induced color change appears to the human eye. The L$_\infty$ norm characterizes the worst-case perturbation magnitude by quantifying the maximum pixel-wise deviation, and we employ the PSNR to assess image fidelity, where higher values indicate reduced noise levels and better preservation of the original visual quality. Finally, we compute SSIM to evaluate the preservation of structural information and luminance consistency, as projector-based perturbations may inadvertently alter local contrast or texture that affects perceived scene geometry, and higher SSIM implies that the adversarial projection better maintains structural fidelity.

\subsection{Results and analysis}
In this section, we present the evaluation results from three perspectives: qualitative comparisons, quantitative assessments based on standardized metrics for attack success and stealthiness, and view ablation studies analyzing the impact of optimization viewpoints on attack performance.

\vspace{.5em}
\noindent\textbf{Qualitative comparisons.}
Representative visualization results are provided in \cref{fig:jersey_untar} and \cref{fig:envelope_tar}, showing the performance of different methods across 20 test views. As shown in \cref{fig:jersey_untar}, our VIPA effectively fools multiple classifiers to predict incorrect labels instead of the original ground-truth class \textbf{jersey}, achieving a greater number of successful misclassifications compared to all the other methods, while simultaneously exhibiting superior stealthiness. Although CAPAA \cite{CAPAA2025} can also generate a single adversarial projection, it is not truly view-invariant, which limits its generalization to unseen viewpoints and results in lower performance compared to our method. In \cref{fig:envelope_tar}, a similar trend is observed, where our method more reliably achieves the target prediction \textbf{file}.

\vspace{.5em}
\noindent\textbf{Quantitative comparisons.} We conduct quantitative comparisons under both untargeted and targeted projector-based attack settings, aggregating results across five real-world setups and evaluating performance on 20 unseen test views per scene. 

In the untargeted attack cases, as presented in \cref{tab:table_compare}, our VIPA attains the highest overall success rate across different thresholds, outperforming all baselines. Importantly, this improvement does not arise from pushing perturbations closer to perceptual limits. Projector-based adversarial attacks often face an inherent trade-off, where stronger attacks typically manifest as larger color shifts, increased noise, or subtle structural distortions that become detectable in camera capture. By contrast, VIPA consistently maintains higher perceptual fidelity across all stealthiness metrics, indicating that its performance improvements stem not from amplifying perturbations, but from generating inherently more effective and visually subtle adversarial projections. This suggests that the adversarial projections better align with classifier-sensitive features that remain stable under viewpoint changes, while avoiding unnecessary modifications in perceptually salient regions.

Compared with the untargeted scenarios, targeted attacks require the adversarial projection to force the classifier into predicting a specific target label rather than simply avoiding the ground truth. Given that targeted attacks are inherently more challenging than untargeted attacks, the performance differences observed in \cref{tab:table_compare} are particularly significant. As shown in the right half of the table, VIPA consistently outperforms all baselines, in line with the qualitative results.

\vspace{.5em}
\noindent\textbf{View ablation.} We conduct a view ablation experiment by varying the number of attack views used during optimization while keeping all other settings unchanged. As shown in \cref{tab:views_ablation}, increasing the number of views consistently improves attack success under both untargeted and targeted settings. When optimizing with only a single view, the generated adversarial projection achieves limited performance. By progressively incorporating more attack views, the adversarial projection becomes increasingly effective across diverse camera perspectives, with the best performance performance achieved using 18 views, indicating that multi-view optimization enhances generalization across unseen viewpoints.

\section{Limitations and Future Work}\label{sec:discussion}

\subsection{Object-related challenges}
Two key object-related factors limit our VIPA’s real-world performance. Stemming from the physical properties of target objects, variations in surface reflectivity and geometric complexity affect how projected patterns are captured, reconstructed, and optimized.

\begin{figure}
\centering
\includegraphics[width=1\linewidth]{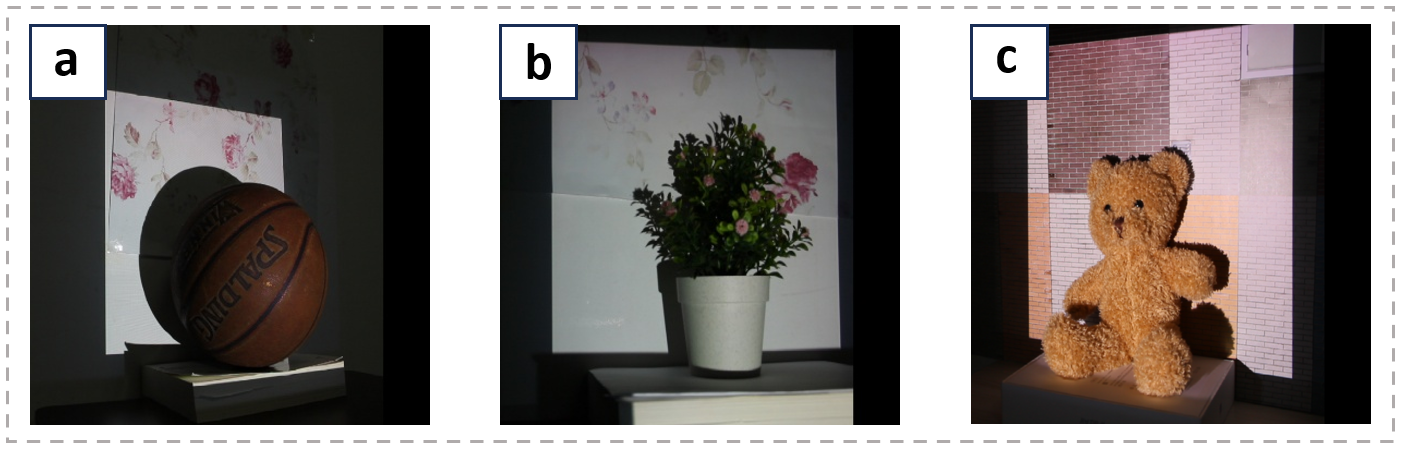}
\caption{Representative failure cases. \textbf{(a)} and \textbf{(b)} correspond to object-related challenges: \textbf{(a)} dark-surfaced basketball and \textbf{(b)} geometrically complex potted plant. \textbf{(c)} illustrates a failed targeted attack case, where all methods failed to misclassify \textbf{teddy} as \textbf{sunscreen} in various test views.}
\label{fig:fails}
\end{figure}

\vspace{.2em}
\noindent\textbf{Dark surface.} Objects with dark surfaces, such as basketballs with deep brown or black regions, as shown in \cref{fig:fails} (a), exhibit significant reconstruction challenges. Due to their low reflectivity, these surfaces absorb a substantial portion of the projected light, leading to inaccurate BRDF shading in the reconstruction. Also, dark regions tend to blend with the object’s own shadows, making contours less distinguishable and resulting in incomplete or imprecise geometry in reconstruction.

\vspace{.2em}
\noindent\textbf{Complex geometry and textures.} Geometrically intricate objects, such as potted plants with dense leaves, as shown in \cref{fig:fails} (b), create severe self-occlusions and strong view-dependent appearance variations. Since our reconstruction pipeline relies on accurately matching the projected calibration patterns across multi-view captures, severe occlusions can lead to incomplete or inconsistent feature correspondences.

In the future, we will try to improve VIPA’s robustness to challenging surfaces and complex geometries. For dark or low-reflectivity objects, adaptive projection or enhanced lighting could improve capture. For intricate shapes, occlusion-aware reconstruction and neural rendering techniques may yield more complete and accurate multi-view correspondences, enhancing real-world performance.

\subsection{System-level constraints and efficiency} 
A practical limitation of VIPA is its sensitivity to object motion during the attack pipeline. Our method assumes that the target object remains stationary throughout calibration, reconstruction, and projection. This assumption is necessary because VIPA relies on a fixed geometric relationship among the projector, cameras, and the object surface to accurately estimate scene geometry and align the optimized perturbation. Even slight unintended movement can break the geometric alignment between the projector, cameras, and object surface, causing the optimized perturbation to be projected inaccurately and often requiring the experiment to be restarted. In addition, the attack is computationally intensive due to 18-view losses per iteration, multi-classifiers, and 30 optimization steps.

These factors limit deployment in uncontrolled or time-sensitive scenarios. Future work will focus on improving motion robustness and reducing computational cost.

\subsection{Low targeted attack success rates}

While VIPA is designed to be classifier-agnostic and view-invariant, targeted attacks remain inherently more challenging than untargeted ones, especially in physical projection scenarios. For example, as shown in \cref{fig:fails} (c), all the methods, including SPAA \cite{SPAA2022}, CAPAA \cite{CAPAA2025} and our methods, fail to misclassify \textbf{teddy} as \textbf{sunscreen}. We observe that such target classes are particularly difficult to attack successfully for all methods, because the original and target categories differ significantly in their visual features and semantic representations. This large visual–semantic gap makes it hard for purely pixel-level perturbations to induce the required shift in high-level features while remaining physically realizable and stealthy under varying viewpoints and illumination.

Future work will focus on improving targeted attacks by bridging the visual and semantic gap between original and target classes, using feature-level guidance and adaptive projection to enhance physical-world success.

\section{Conclusion}\label{sec:conclusion}

In this paper, we proposed VIPA, a projector-based adversarial attack method that is both view-invariant and classifier-agnostic. VIPA applies a view-invariant project-and-capture simulation model and jointly optimizes losses across multiple views and diverse classifiers. This enables the generation of a single adversarial pattern that, when projected onto a scene, consistently causes misclassification across classifiers and views, including variations in horizontal and vertical perspectives. Experimental results demonstrate that our method achieves superior success rates and stealthiness compared to existing approaches, in both targeted and untargeted attack scenarios.

\bibliographystyle{abbrv-doi-hyperref}
\bibliography{ref}

\clearpage

\appendix

\setcounter{equation}{0}
\setcounter{figure}{0}
\setcounter{table}{0}

\supptitle{VIPA: View-Invariant Projector-Based Adversarial Attack}

\makeMySupplementaryTitle
\def\eg{\emph{e.g}.} \def\Eg{\emph{E.g}.}
\def\ie{\emph{i.e}.} \def\Ie{\emph{I.e}.}
\def\cf{\emph{c.f}.} \def\Cf{\emph{C.f}.}
\def\etc{\emph{etc}.} \def\vs{\emph{vs}.}
\def\wrt{w.r.t.} \def\dof{d.o.f.}
\def\etal{\emph{et al}.}

\renewcommand{\figureautorefname}{Fig.}
\renewcommand{\tableautorefname}{Table} 
\renewcommand{\equationautorefname}{Eqn.} 
\renewcommand{\sectionautorefname}{\S} 
\renewcommand{\subsectionautorefname}{\S} 

\section{Introduction}
In this supplementary material, we provide additional experimental results, including more data for untargeted and targeted attack to further validate the effectiveness and robustness of our proposed VIPA method. We show the results under five different setups. For each setup, we include a quantitative comparison table covering both untargeted and targeted attacks with baseline methods for reference, and a qualitative comparison figure that visualizes the attack outcomes of three views alongside the same baselines. Note that, we randomly sample 20 test views for each setup, and evaluate three \(d_{\text{thr}}\) values and three classifiers, resulting in a total of 180 attack trials per setup for untargeted attacks and another 180 trials for targeted attacks.
\vspace{1.5cm}

\begin{figure*}[!htbp]
	\begin{center}
		\includegraphics[width=1\linewidth]{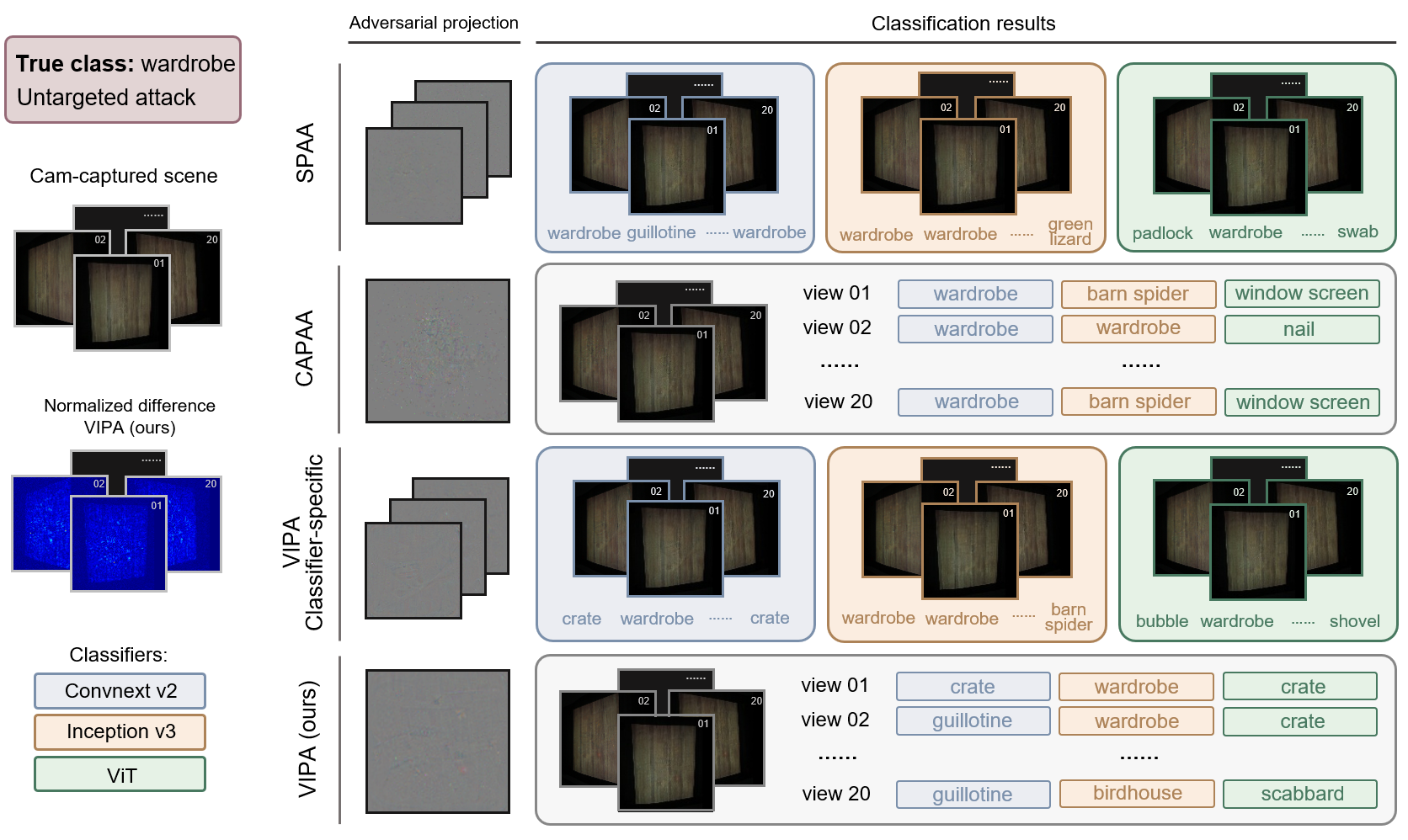}
		\caption{
        \textbf{Untargeted} projector-based adversarial attack. The goal is to use adversarial projections to induce various classifiers to misclassify the camera-captured scene as any label other than the true class, \textbf{wardrobe}, from various views. For evaluation, we randomly sample and capture 20 test viewpoints. On the far left, we present the clean camera-captured scene and the corresponding normalized difference maps computed against the camera-captured images obtained after projecting the adversarial pattern produced by our method. For each approach, the 1\textsuperscript{st} column visualizes its adversarial projections. Note that CAPAA \cite{CAPAA2025} and VIPA (ours) are classifier-agnostic and therefore produce a single adversarial pattern for all classifiers, whereas VIPA classifier-specific and SPAA \cite{SPAA2022} generate distinct adversarial patterns for each of the three classifiers. In the classification results, we consider three classifiers, Convnext v2 \cite{woo2023convnextv2}, Inception v3 \cite{2016inceptionv3} and ViT \cite{2021ViT}, which are denoted in \protect\colorboxblock{cblue_border}{cblue_fill},  \protect\colorboxblock{corange_border}{corange_fill} and \protect\colorboxblock{cgreen_border}{cgreen_fill}, respectively, and annotate each classifier’s predicted label next to the corresponding viewpoint.
        }\label{fig:wardrobe_untar}
	\end{center}
\end{figure*}

\clearpage
\begin{figure*}[t]
	\begin{center}
		\includegraphics[width=1\linewidth]{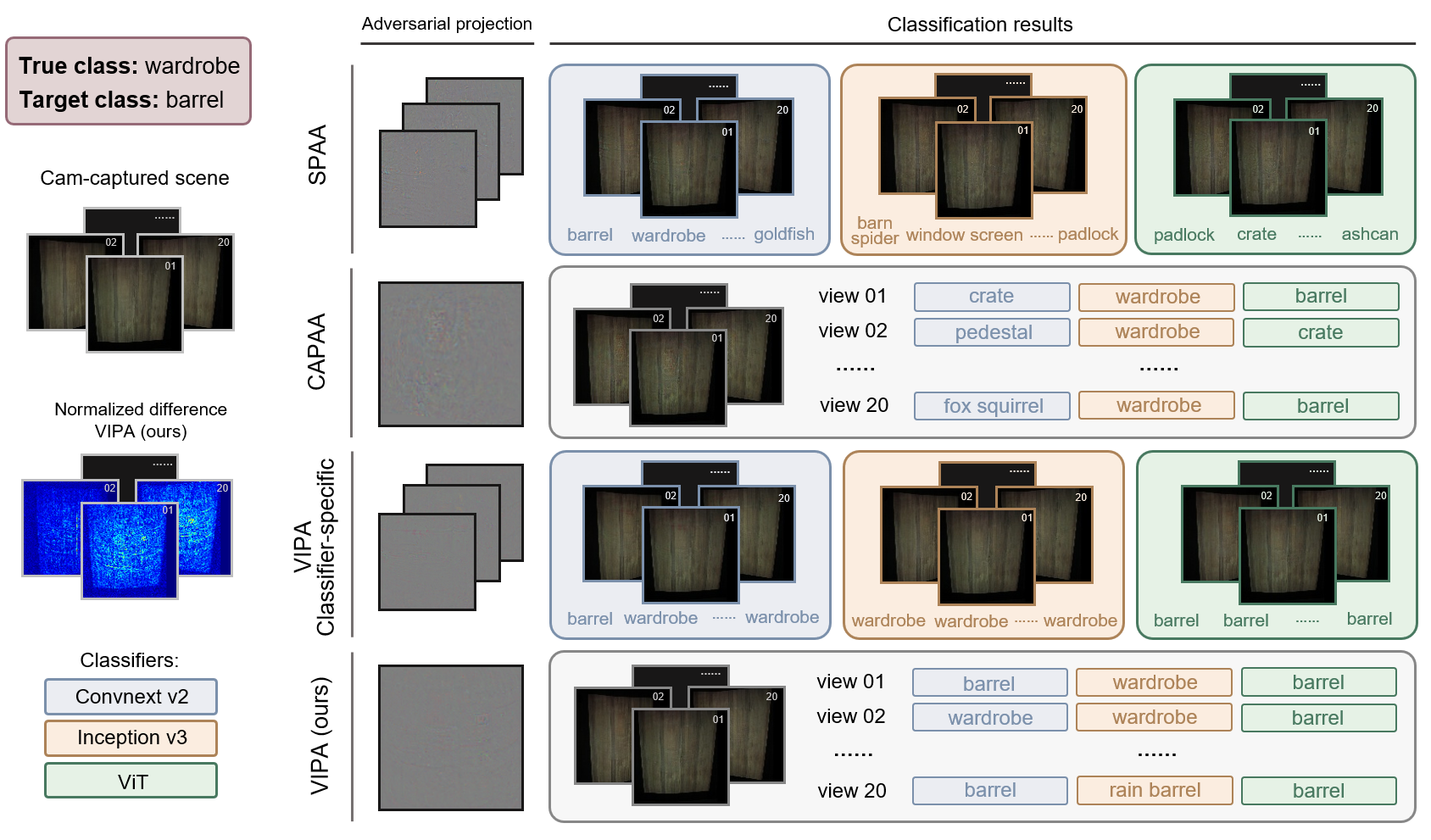}
		\caption{
        \textbf{Targeted} projector-based adversarial attack. The goal is to use adversarial projections to cause different classifiers to misclassify the camera-captured scene \textbf{wardrobe} as \textbf{barrel} from various viewpoints. For evaluation, we randomly sample and capture 20 test viewpoints. On the far left, we present the clean camera-captured scene and the corresponding normalized difference maps computed against the camera-captured images obtained after projecting the adversarial pattern produced by our method. For each approach, the 1\textsuperscript{st} column visualizes its adversarial projections. Note that CAPAA \cite{CAPAA2025} and VIPA (ours) are classifier-agnostic and therefore produce a single adversarial pattern for all classifiers, whereas VIPA classifier-specific and SPAA \cite{SPAA2022} generate distinct adversarial patterns for each of the three classifiers. In the classification results, we consider three classifiers, Convnext v2 \cite{woo2023convnextv2}, Inception v3 \cite{2016inceptionv3} and ViT \cite{2021ViT}, which are denoted in \protect\colorboxblock{cblue_border}{cblue_fill},  \protect\colorboxblock{corange_border}{corange_fill} and \protect\colorboxblock{cgreen_border}{cgreen_fill}, respectively, and annotate each classifier’s predicted label next to the corresponding viewpoint.
        }\label{fig:wardrobe_tar}
	\end{center}
\end{figure*}
\begin{table*}[!t]
\centering
\fontsize{8.1pt}{11pt}\selectfont
\caption{Quantitative comparison of different attack methods under untargeted and targeted attack scenarios. For untargeted attacks, adversarial projections are designed to cause camera-captured instances of the true class \textbf{wardrobe} to be misclassified as any other class. For targeted attacks, the objective is to consistently fool classifiers into predicting \textbf{barrel} for scenes whose true label is \textbf{wardrobe}. \textbf{OSR} is the success rate for corresponding rows.}
\label{tab:table_compare_wardrobe}
\resizebox{\linewidth}{!}{
\begin{tabular}{@{}lcccccccccccccc@{}}
    \toprule[0.5mm]
    \multirow{2}{*}{\textbf{Method}} & \multirow{2}{*}{\textbf{d$_\textbf{thr}$}} 
    & \multicolumn{6}{c}{\textbf{Untargeted attack}} 
    & \multicolumn{6}{c}{\textbf{Targeted attack}} \\ 
    \cmidrule(lr){3-8} \cmidrule(lr){9-14}
     &  & \textbf{$\Delta$E$\downarrow$} & \textbf{L$_\infty\downarrow$} & \textbf{PSNR$\uparrow$} & \textbf{SSIM$\uparrow$} & \textbf{OSR$\uparrow$} & \textbf{Avg. OSR$\uparrow$} 
     & \textbf{$\Delta$E$\downarrow$} & \textbf{L$_\infty\downarrow$} & \textbf{PSNR$\uparrow$} & \textbf{SSIM$\uparrow$} & \textbf{OSR$\uparrow$} & \textbf{Avg. OSR$\uparrow$} \\ 
    \midrule
    \multirow{3}{*}{SPAA \cite{SPAA2022}} 
     & 4 & 1.40 & 2.14 & 41.83 & 0.9713 & 48.33\% & \multirow{3}{*}{52.78\%}
         & 1.66 & 2.71 & 39.56 & 0.9641 & 8.33\% & \multirow{3}{*}{4.44\%} \\
     & 5 & 1.42 & 2.22 & 41.33 & 0.9684 & 43.33\% &  
         & 1.79 & 3.01 & 38.39 & 0.9582 & 1.67\% &  \\
     & 6 & 1.47 & 2.30 & 40.76 & 0.9657 & 66.67\% &  
         & 2.03 & 3.34 & 37.44 & 0.9529 & 3.33\% &  \\
    \cmidrule[0.3mm]{1-14}
    \multirow{3}{*}{CAPAA \cite{CAPAA2025}} 
     & 4 & 1.48 & 2.30 & 40.73 & 0.9662 & 45.00\% & \multirow{3}{*}{49.44\%}
         & 1.84 & 2.93 & 38.93 & 0.9615 & 11.67\% & \multirow{3}{*}{11.11\%} \\
     & 5 & 1.48 & 2.28 & 40.81 & 0.9658 & 55.00\% &  
         & 1.83 & 2.97 & 38.83 & 0.9626 & 16.67\% &  \\
     & 6 & 1.45 & 2.27 & 40.76 & 0.9659 & 48.33\% &  
         & 1.83 & 2.92 & 38.91 & 0.9612 & 5.00\% &  \\
    \cmidrule[0.3mm]{1-14}
    \multirow{3}{*}{\shortstack[l]{VIPA \\classifier-specific}}  
     & 4 & 1.36 & 2.25 & 41.53 & 0.9736 & 65.00\% & \multirow{3}{*}{63.89\%}
         & 1.49 & 2.34 & 41.75 & 0.9764 & 16.67\% & \multirow{3}{*}{11.11\%} \\
     & 5 & 1.36 & 2.22 & 41.64 & 0.9737 & 60.00\% &  
         & 1.47 & 2.32 & 41.79 & 0.9764 & 8.33\% &  \\
     & 6 & 1.34 & 2.21 & 41.63 & 0.9733 & 66.67\% &  
         & 1.47 & 2.35 & 41.69 & 0.9763 & 8.33\% &  \\
    \cmidrule[0.3mm]{1-14}
    \multirow{3}{*}{VIPA (ours)} 
     & 4 & 1.42 & 2.33 & 41.13 & 0.9714 & 66.67\% & \multirow{3}{*}{\textbf{70.56\%}}
         & 1.48 & 2.34 & 41.65 & 0.9757 & 25.00\% & \multirow{3}{*}{\textbf{18.33\%}} \\
     & 5 & 1.38 & 2.27 & 41.33 & 0.9726 & 75.00\% &  
         & 1.47 & 2.34 & 41.64 & 0.9760 & 13.33\% &  \\
     & 6 & 1.40 & 2.27 & 41.45 & 0.9733 & 70.00\% &  
         & 1.47 & 2.27 & 41.95 & 0.9763 & 16.67\% &  \\
    \bottomrule[0.5mm]
\end{tabular}}
\end{table*}

\clearpage
\begin{figure*}[t]
	\begin{center}
		\includegraphics[width=1\linewidth]{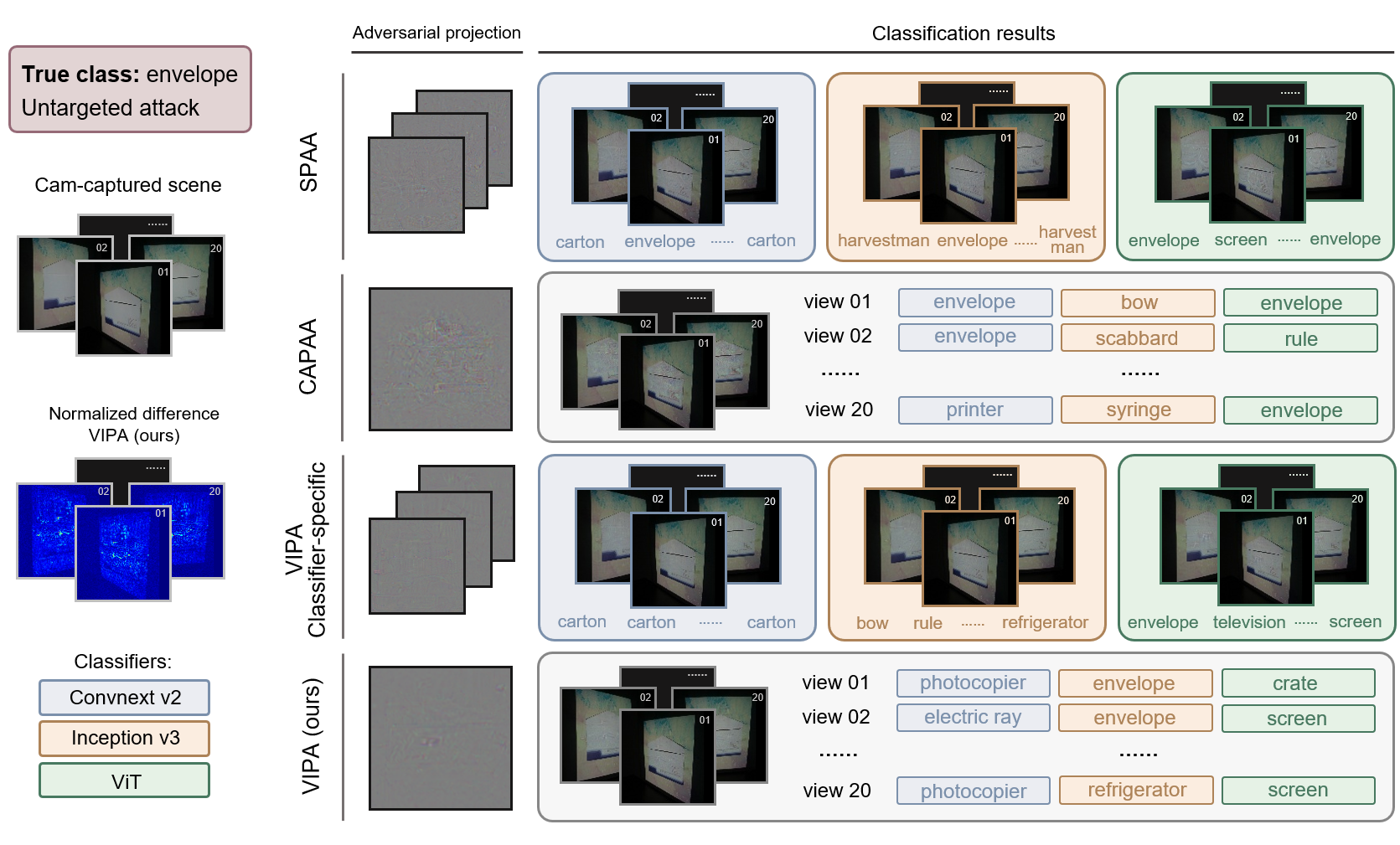}
		\caption{
        \textbf{Untargeted} projector-based adversarial attack. The goal is to use adversarial projections to induce various classifiers to misclassify the camera-captured scene as any label other than the true class, \textbf{envelope}, from various views. For evaluation, we randomly sample and capture 20 test viewpoints. On the far left, we present the clean camera-captured scene and the corresponding normalized difference maps computed against the camera-captured images obtained after projecting the adversarial pattern produced by our method. For each approach, the 1\textsuperscript{st} column visualizes its adversarial projections. Note that CAPAA \cite{CAPAA2025} and VIPA (ours) are classifier-agnostic and therefore produce a single adversarial pattern for all classifiers, whereas VIPA classifier-specific and SPAA \cite{SPAA2022} generate distinct adversarial patterns for each of the three classifiers. In the classification results, we consider three classifiers, Convnext v2 \cite{woo2023convnextv2}, Inception v3 \cite{2016inceptionv3} and ViT \cite{2021ViT}, which are denoted in \protect\colorboxblock{cblue_border}{cblue_fill},  \protect\colorboxblock{corange_border}{corange_fill} and \protect\colorboxblock{cgreen_border}{cgreen_fill}, respectively, and annotate each classifier’s predicted label next to the corresponding viewpoint.
        }\label{fig:envelope_untar}
	\end{center}
\end{figure*}
\begin{table*}[!t]
\centering
\fontsize{8.1pt}{11pt}\selectfont
\caption{Quantitative comparison of different attack methods under untargeted and targeted attack scenarios. For untargeted attacks, adversarial projections are designed to cause camera-captured instances of the true class \textbf{envelope} to be misclassified as any other class. For targeted attacks, the objective is to consistently fool classifiers into predicting \textbf{file} for scenes whose true label is \textbf{envelope}. \textbf{OSR} is the success rate for corresponding rows.}
\label{tab:table_compare_envelope}
\resizebox{\linewidth}{!}{
\begin{tabular}{@{}lcccccccccccccc@{}}
    \toprule[0.5mm]
    \multirow{2}{*}{\textbf{Method}} & \multirow{2}{*}{\textbf{d$_\textbf{thr}$}} 
    & \multicolumn{6}{c}{\textbf{Untargeted attack}} 
    & \multicolumn{6}{c}{\textbf{Targeted attack}} \\ 
    \cmidrule(lr){3-8} \cmidrule(lr){9-14}
     &  & \textbf{$\Delta$E$\downarrow$} & \textbf{L$_\infty\downarrow$} & \textbf{PSNR$\uparrow$} & \textbf{SSIM$\uparrow$} & \textbf{OSR$\uparrow$} & \textbf{Avg. OSR$\uparrow$} 
     & \textbf{$\Delta$E$\downarrow$} & \textbf{L$_\infty\downarrow$} & \textbf{PSNR$\uparrow$} & \textbf{SSIM$\uparrow$} & \textbf{OSR$\uparrow$} & \textbf{Avg. OSR$\uparrow$} \\ 
    \midrule
    \multirow{3}{*}{SPAA \cite{SPAA2022}}
     & 4 & 1.73 & 3.03 & 38.24 & 0.9630 & 61.67\% & \multirow{3}{*}{62.78\%}
         & 1.69 & 3.03 & 38.10 & 0.9624 & 10.00\% & \multirow{3}{*}{7.78\%} \\
     & 5 & 1.89 & 3.47 & 36.73 & 0.9553 & 63.33\% &  
         & 1.88 & 3.46 & 36.79 & 0.9545 & 1.67\% &  \\
     & 6 & 2.13 & 3.99 & 35.47 & 0.9455 & 63.33\% &
         & 2.13 & 3.99 & 35.50 & 0.9461 & 11.67\% &  \\
    \cmidrule[0.3mm]{1-14}
    \multirow{3}{*}{CAPAA \cite{CAPAA2025}}
     & 4 & 2.06 & 3.79 & 35.98 & 0.9543 & 63.33\% & \multirow{3}{*}{65.00\%}
         & 2.06 & 3.71 & 36.43 & 0.9557 & 20.00\% & \multirow{3}{*}{17.22\%} \\
     & 5 & 2.05 & 3.78 & 35.87 & 0.9529 & 55.00\% &
         & 1.98 & 3.58 & 36.61 & 0.9559 & 16.67\% &  \\
     & 6 & 2.08 & 3.79 & 35.94 & 0.9532 & 76.67\% &
         & 2.01 & 3.62 & 36.54 & 0.9560 & 15.00\% &  \\
    \cmidrule[0.3mm]{1-14}
    \multirow{3}{*}{\shortstack[l]{VIPA \\classifier-specific}}
     & 4 & 1.61 & 2.92 & 39.30 & 0.9754 & 66.67\% & \multirow{3}{*}{69.44\%}
         & 1.59 & 2.91 & 39.34 & 0.9765 & 11.67\% & \multirow{3}{*}{12.22\%} \\
     & 5 & 1.60 & 2.92 & 39.28 & 0.9754 & 71.67\% & 
         & 1.59 & 2.92 & 39.31 & 0.9765 & 20.00\% &  \\
     & 6 & 1.60 & 2.93 & 39.26 & 0.9753 & 70.00\% &
         & 1.62 & 2.93 & 39.29 & 0.9764 & 5.00\% &  \\
    \cmidrule[0.3mm]{1-14}
    \multirow{3}{*}{VIPA (ours)}
     & 4 & 1.63 & 2.96 & 39.12 & 0.9748 & 71.67\% & \multirow{3}{*}{\textbf{76.11\%}}
         & 1.67 & 3.01 & 38.98 & 0.9763 & 25.00\% & \multirow{3}{*}{\textbf{28.33\%}} \\
     & 5 & 1.65 & 2.98 & 39.04 & 0.9743 & 70.00\% &
         & 1.59 & 2.94 & 39.16 & 0.9761 & 20.00\% &  \\
     & 6 & 1.64 & 2.97 & 39.01 & 0.9742 & 86.67\% &
         & 1.61 & 2.90 & 39.34 & 0.9762 & 40.00\% &  \\
    \bottomrule[0.5mm]
\end{tabular}
}
\end{table*}

\clearpage
\begin{figure*}[t]
	\begin{center}
		\includegraphics[width=1\linewidth]{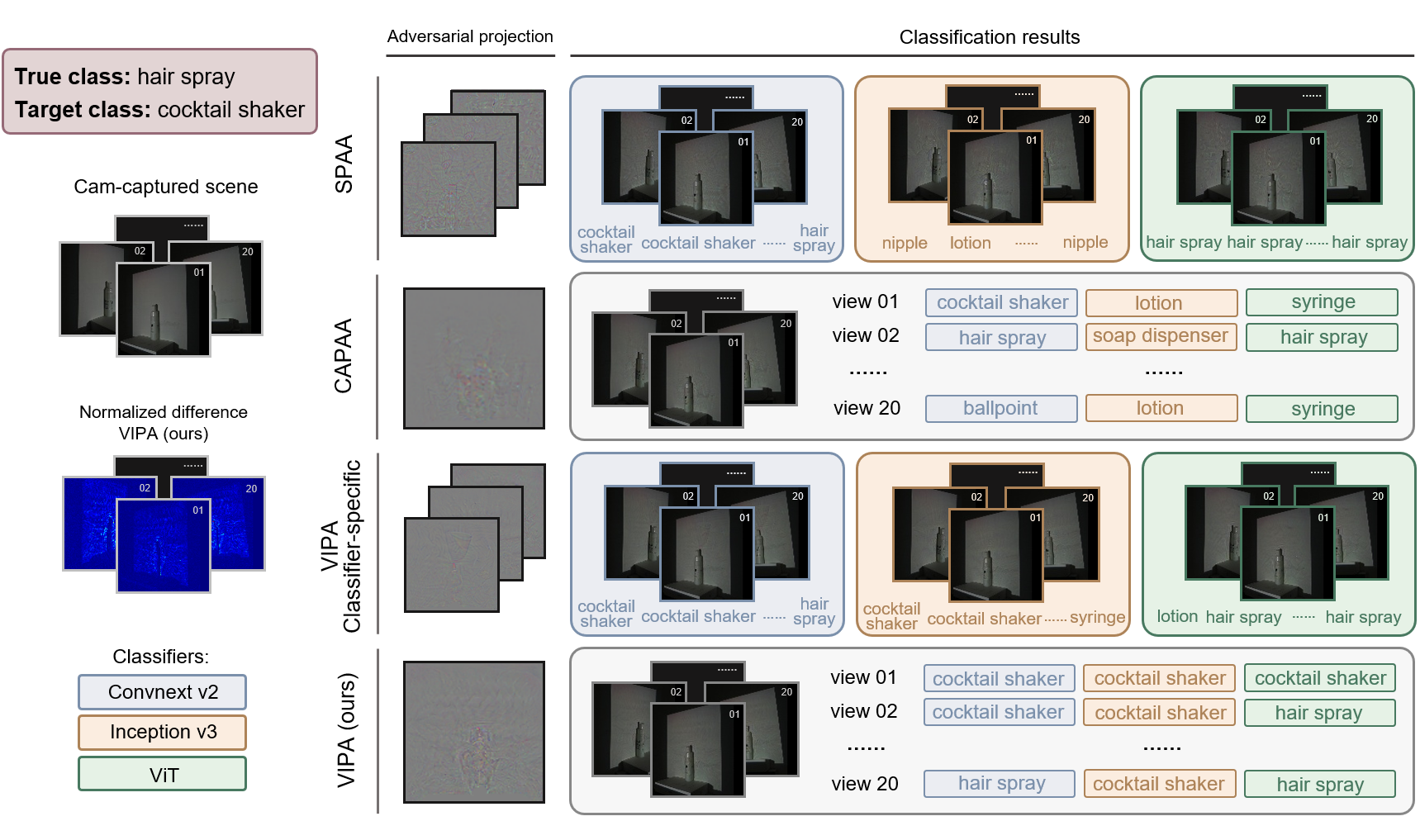}
		\caption{
        \textbf{Targeted} projector-based adversarial attack. The goal is to use adversarial projections to cause different classifiers to misclassify the camera-captured scene \textbf{hair spray} as \textbf{cocktail shaker} from various viewpoints. For evaluation, we randomly sample and capture 20 test viewpoints. On the far left, we present the clean camera-captured scene and the corresponding normalized difference maps computed against the camera-captured images obtained after projecting the adversarial pattern produced by our method. For each approach, the 1\textsuperscript{st} column visualizes its adversarial projections. Note that CAPAA \cite{CAPAA2025} and VIPA (ours) are classifier-agnostic and therefore produce a single adversarial pattern for all classifiers, whereas VIPA classifier-specific and SPAA \cite{SPAA2022} generate distinct adversarial patterns for each of the three classifiers. In the classification results, we consider three classifiers, Convnext v2 \cite{woo2023convnextv2}, Inception v3 \cite{2016inceptionv3} and ViT \cite{2021ViT}, which are denoted in \protect\colorboxblock{cblue_border}{cblue_fill},  \protect\colorboxblock{corange_border}{corange_fill} and \protect\colorboxblock{cgreen_border}{cgreen_fill}, respectively, and annotate each classifier’s predicted label next to the corresponding viewpoint.
        }\label{fig:hairspray_tar}
	\end{center}
\end{figure*}
\begin{table*}[!t]
\centering
\fontsize{8.1pt}{11pt}\selectfont
\caption{Quantitative comparison of different attack methods under untargeted and targeted attack scenarios. For untargeted attacks, adversarial projections are designed to cause camera-captured instances of the true class \textbf{hair spray} to be misclassified as any other class. For targeted attacks, the objective is to consistently fool classifiers into predicting \textbf{cocktail shaker} for scenes whose true label is \textbf{hair spray}. \textbf{OSR} is the success rate for corresponding rows.}
\label{tab:table_compare_hairspray}
\resizebox{\linewidth}{!}{
\begin{tabular}{@{}lcccccccccccccc@{}}
    \toprule[0.5mm]
    \multirow{2}{*}{\textbf{Method}} & \multirow{2}{*}{\textbf{d$_\textbf{thr}$}} 
    & \multicolumn{6}{c}{\textbf{Untargeted attack}} 
    & \multicolumn{6}{c}{\textbf{Targeted attack}} \\ 
    \cmidrule(lr){3-8} \cmidrule(lr){9-14}
     &  & \textbf{$\Delta$E$\downarrow$} & \textbf{L$_\infty\downarrow$} & \textbf{PSNR$\uparrow$} & \textbf{SSIM$\uparrow$} & \textbf{OSR$\uparrow$} & \textbf{Avg. OSR$\uparrow$} 
     & \textbf{$\Delta$E$\downarrow$} & \textbf{L$_\infty\downarrow$} & \textbf{PSNR$\uparrow$} & \textbf{SSIM$\uparrow$} & \textbf{OSR$\uparrow$} & \textbf{Avg. OSR$\uparrow$} \\ 
    \midrule
    \multirow{3}{*}{SPAA \cite{SPAA2022}}
     & 4 & 1.50 & 2.54 & 39.53 & 0.9689 & 53.33\% & \multirow{3}{*}{46.67\%}
         & 1.49 & 2.51 & 39.55 & 0.9698 & 11.67\% & \multirow{3}{*}{12.78\%} \\
     & 5 & 1.74 & 2.92 & 38.13 & 0.9615 & 46.67\% &  
         & 1.74 & 2.91 & 38.09 & 0.9622 & 10.00\% &  \\
     & 6 & 2.10 & 3.32 & 37.12 & 0.9551 & 40.00\% &
         & 2.06 & 3.28 & 37.17 & 0.9566 & 16.67\% &  \\
    \cmidrule[0.3mm]{1-14}
    \multirow{3}{*}{CAPAA \cite{CAPAA2025}}
     & 4 & 1.43 & 2.34 & 40.24 & 0.9753 & 60.00\% & \multirow{3}{*}{59.44\%}
         & 1.46 & 2.38 & 40.31 & 0.9761 & 26.67\% & \multirow{3}{*}{23.89\%} \\
     & 5 & 1.43 & 2.31 & 40.49 & 0.9756 & 58.33\% &
         & 1.46 & 2.38 & 40.43 & 0.9762 & 20.00\% &  \\
     & 6 & 1.44 & 2.34 & 40.42 & 0.9762 & 60.00\% &
         & 1.44 & 2.33 & 40.62 & 0.9769 & 25.00\% &  \\
    \cmidrule[0.3mm]{1-14}
    \multirow{3}{*}{\shortstack[l]{VIPA \\classifier-specific}}
     & 4 & 1.48 & 2.40 & 40.60 & 0.9762 & 63.33\% & \multirow{3}{*}{58.89\%}
         & 1.49 & 2.38 & 40.75 & 0.9774 & 25.00\% & \multirow{3}{*}{23.89\%} \\
     & 5 & 1.49 & 2.41 & 40.59 & 0.9762 & 56.67\% & 
         & 1.50 & 2.38 & 40.77 & 0.9773 & 21.67\% &  \\
     & 6 & 1.47 & 2.38 & 40.66 & 0.9763 & 56.67\% &
         & 1.50 & 2.40 & 40.72 & 0.9773 & 25.00\% &  \\
    \cmidrule[0.3mm]{1-14}
    \multirow{3}{*}{VIPA (ours)}
     & 4 & 1.50 & 2.44 & 40.37 & 0.9744 & 58.33\% & \multirow{3}{*}{\textbf{63.33\%}}
         & 1.55 & 2.47 & 40.26 & 0.9741 & 31.67\% & \multirow{3}{*}{\textbf{40.00\%}} \\
     & 5 & 1.51 & 2.44 & 40.35 & 0.9745 & 65.00\% &
         & 1.55 & 2.47 & 40.24 & 0.9742 & 45.00\% &  \\
     & 6 & 1.52 & 2.45 & 40.35 & 0.9743 & 66.67\% &
         & 1.58 & 2.50 & 40.17 & 0.9739 & 43.33\% &  \\
    \bottomrule[0.5mm]
\end{tabular}
}
\end{table*}

\clearpage
\begin{figure*}[t]
	\begin{center}
		\includegraphics[width=1\linewidth]{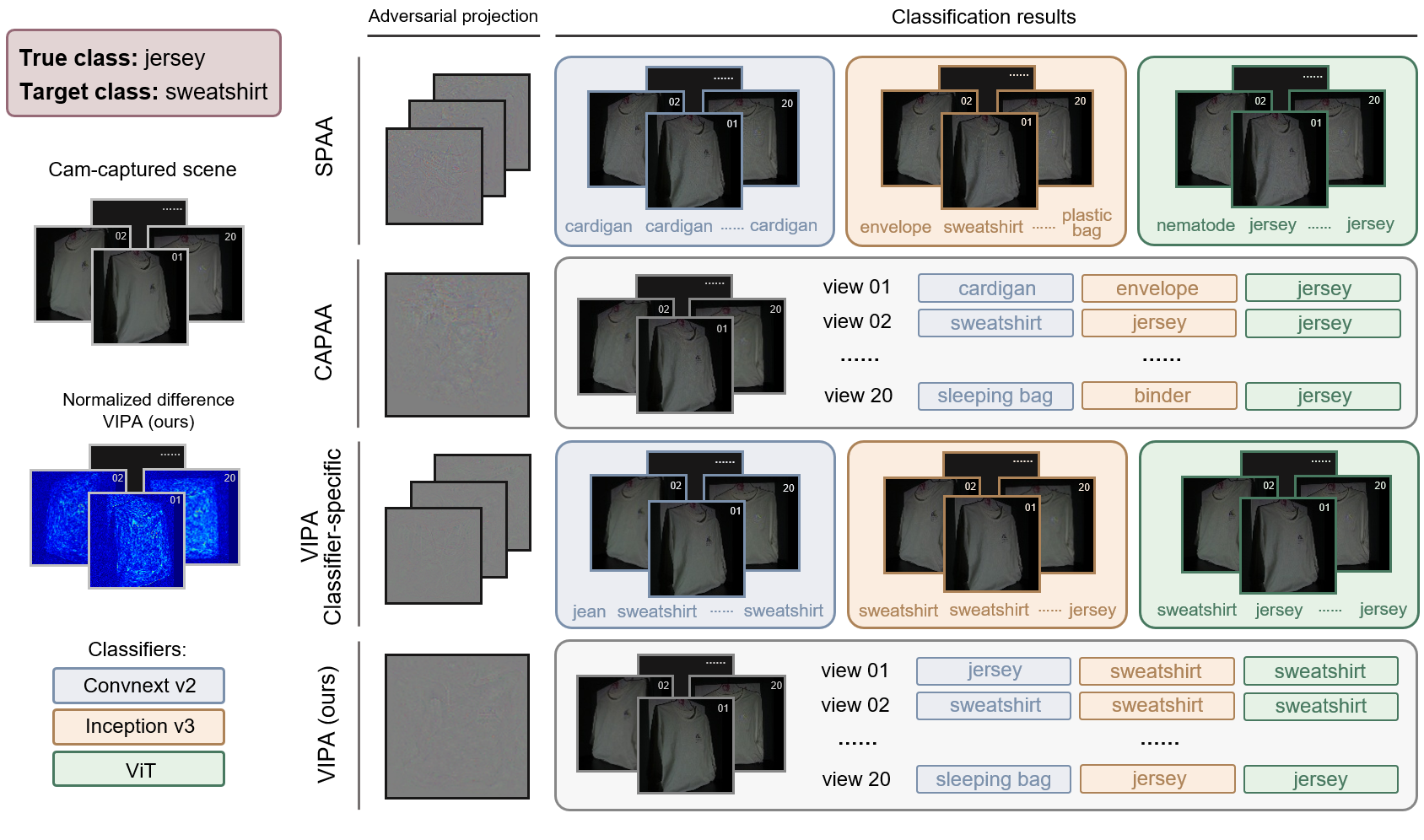}
		\caption{
        \textbf{Targeted} projector-based adversarial attack. The goal is to use adversarial projections to cause different classifiers to misclassify the camera-captured scene \textbf{jersey} as \textbf{sweatshirt} from various viewpoints. For evaluation, we randomly sample and capture 20 test viewpoints. On the far left, we present the clean camera-captured scene and the corresponding normalized difference maps computed against the camera-captured images obtained after projecting the adversarial pattern produced by our method. For each approach, the 1\textsuperscript{st} column visualizes its adversarial projections. Note that CAPAA \cite{CAPAA2025} and VIPA (ours) are classifier-agnostic and therefore produce a single adversarial pattern for all classifiers, whereas VIPA classifier-specific and SPAA \cite{SPAA2022} generate distinct adversarial patterns for each of the three classifiers. In the classification results, we consider three classifiers, Convnext v2 \cite{woo2023convnextv2}, Inception v3 \cite{2016inceptionv3} and ViT \cite{2021ViT}, which are denoted in \protect\colorboxblock{cblue_border}{cblue_fill},  \protect\colorboxblock{corange_border}{corange_fill} and \protect\colorboxblock{cgreen_border}{cgreen_fill}, respectively, and annotate each classifier’s predicted label next to the corresponding viewpoint.
        }\label{fig:jersey_tar}
	\end{center}
\end{figure*}
\begin{table*}[!t]
\centering
\fontsize{8.1pt}{11pt}\selectfont
\caption{Quantitative comparison of different attack methods under untargeted and targeted attack scenarios. For untargeted attacks, adversarial projections are designed to cause camera-captured instances of the true class \textbf{jersey} to be misclassified as any other class. For targeted attacks, the objective is to consistently fool classifiers into predicting \textbf{sweatshirt} for scenes whose true label is \textbf{jersey}. \textbf{OSR} is the success rate for corresponding rows.}
\label{tab:table_compare_jersey}
\resizebox{\linewidth}{!}{
\begin{tabular}{@{}lcccccccccccccc@{}}
    \toprule[0.5mm]
    \multirow{2}{*}{\textbf{Method}} & \multirow{2}{*}{\textbf{d$_\textbf{thr}$}} 
    & \multicolumn{6}{c}{\textbf{Untargeted attack}} 
    & \multicolumn{6}{c}{\textbf{Targeted attack}} \\ 
    \cmidrule(lr){3-8} \cmidrule(lr){9-14}
     &  & \textbf{$\Delta$E$\downarrow$} & \textbf{L$_\infty\downarrow$} & \textbf{PSNR$\uparrow$} & \textbf{SSIM$\uparrow$} & \textbf{OSR$\uparrow$} & \textbf{Avg. OSR$\uparrow$} 
     & \textbf{$\Delta$E$\downarrow$} & \textbf{L$_\infty\downarrow$} & \textbf{PSNR$\uparrow$} & \textbf{SSIM$\uparrow$} & \textbf{OSR$\uparrow$} & \textbf{Avg. OSR$\uparrow$} \\ 
    \midrule
    \multirow{3}{*}{SPAA \cite{SPAA2022}}
     & 4 & 1.73 & 2.74 & 39.66 & 0.9713 & 73.33\% & \multirow{3}{*}{70.00\%}
         & 1.54 & 2.47 & 40.54 & 0.9731 & 1.67\% & \multirow{3}{*}{5.00\%} \\
     & 5 & 1.78 & 2.98 & 38.64 & 0.9661 & 63.33\% &  
         & 1.65 & 2.71 & 39.42 & 0.9681 & 6.67\% &  \\
     & 6 & 1.97 & 3.32 & 37.55 & 0.9591 & 73.33\% &
         & 1.79 & 2.97 & 38.48 & 0.9626 & 6.67\% &  \\
    \cmidrule[0.3mm]{1-14}
    \multirow{3}{*}{CAPAA \cite{CAPAA2025}}
     & 4 & 1.67 & 2.85 & 39.27 & 0.9726 & 45.00\% & \multirow{3}{*}{53.89\%}
         & 1.50 & 2.51 & 40.47 & 0.9744 & 45.00\% & \multirow{3}{*}{36.11\%} \\
     & 5 & 1.66 & 2.80 & 39.48 & 0.9726 & 63.33\% &
         & 1.51 & 2.55 & 40.30 & 0.9744 & 30.00\% &  \\
     & 6 & 1.65 & 2.82 & 39.34 & 0.9723 & 53.33\% &
         & 1.54 & 2.57 & 40.26 & 0.9745 & 33.33\% &  \\
    \cmidrule[0.3mm]{1-14}
    \multirow{3}{*}{\shortstack[l]{VIPA \\classifier-specific}}
     & 4 & 1.46 & 2.51 & 40.74 & 0.9798 & 78.33\% & \multirow{3}{*}{78.33\%}
         & 1.33 & 2.22 & 41.96 & 0.9817 & 56.67\% & \multirow{3}{*}{50.56\%} \\
     & 5 & 1.49 & 2.55 & 40.60 & 0.9798 & 73.33\% & 
         & 1.35 & 2.27 & 41.74 & 0.9818 & 50.00\% &  \\
     & 6 & 1.46 & 2.51 & 40.73 & 0.9798 & 83.33\% &
         & 1.34 & 2.25 & 41.84 & 0.9817 & 45.00\% &  \\
    \cmidrule[0.3mm]{1-14}
    \multirow{3}{*}{VIPA (ours)}
     & 4 & 1.48 & 2.51 & 40.73 & 0.9798 & 90.00\% & \multirow{3}{*}{\textbf{91.11\%}}
         & 1.34 & 2.23 & 41.88 & 0.9818 & 60.00\% & \multirow{3}{*}{\textbf{55.56\%}} \\
     & 5 & 1.50 & 2.57 & 40.49 & 0.9793 & 96.67\% &
         & 1.39 & 2.31 & 41.60 & 0.9817 & 51.67\% &  \\
     & 6 & 1.54 & 2.62 & 40.40 & 0.9795 & 86.67\% &
         & 1.33 & 2.26 & 41.71 & 0.9817 & 55.00\% &  \\
    \bottomrule[0.5mm]
\end{tabular}
}
\end{table*}

\clearpage
\begin{figure*}[t]
	\begin{center}
		\includegraphics[width=1\linewidth]{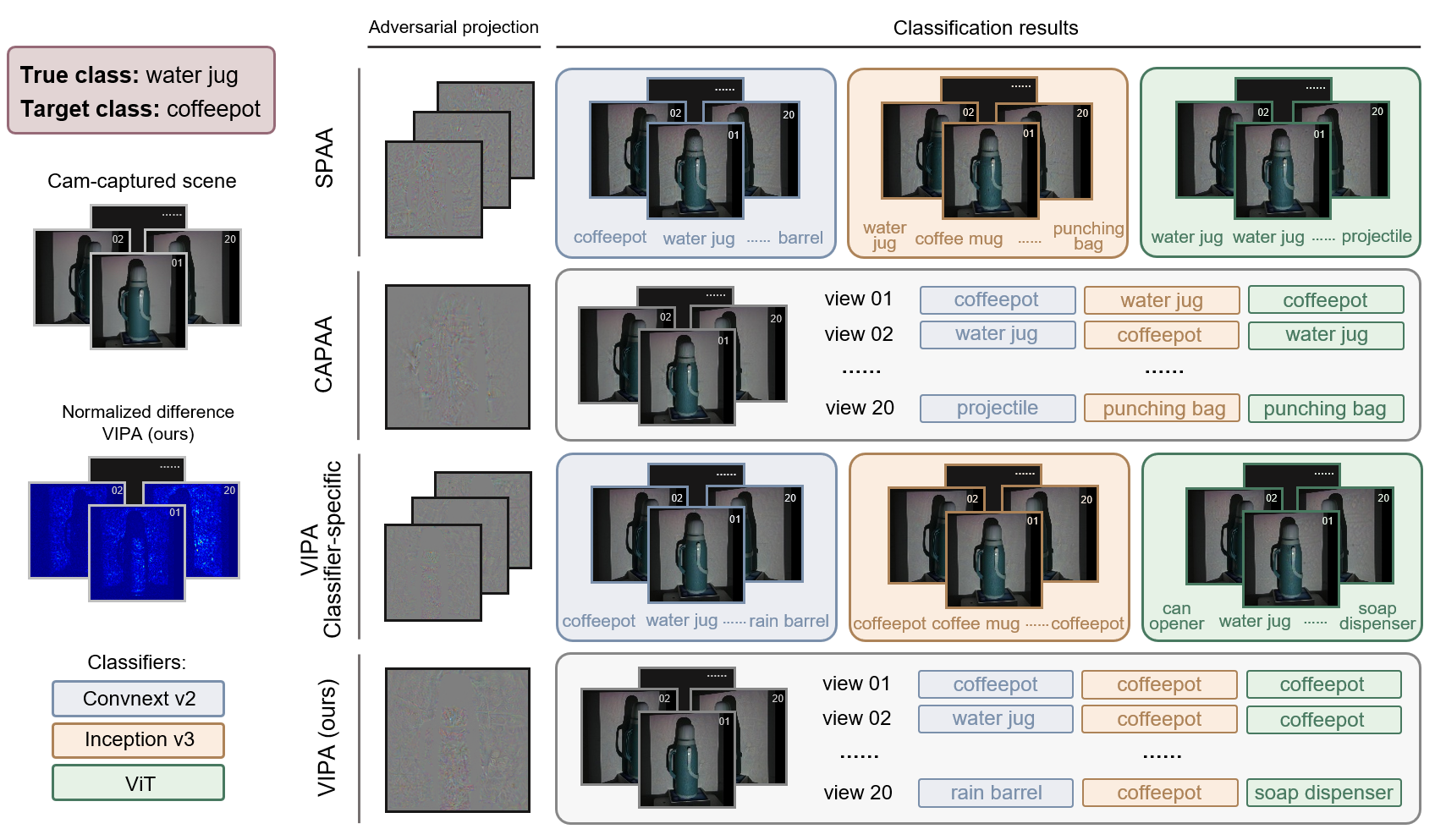}
		\caption{
        \textbf{Untargeted} projector-based adversarial attack. The goal is to use adversarial projections to induce various classifiers to misclassify the camera-captured scene as any label other than the true class, \textbf{water jug}, from various views. For evaluation, we randomly sample and capture 20 test viewpoints. On the far left, we present the clean camera-captured scene and the corresponding normalized difference maps computed against the camera-captured images obtained after projecting the adversarial pattern produced by our method. For each approach, the 1\textsuperscript{st} column visualizes its adversarial projections. Note that CAPAA \cite{CAPAA2025} and VIPA (ours) are classifier-agnostic and therefore produce a single adversarial pattern for all classifiers, whereas VIPA classifier-specific and SPAA \cite{SPAA2022} generate distinct adversarial patterns for each of the three classifiers. In the classification results, we consider three classifiers, Convnext v2 \cite{woo2023convnextv2}, Inception v3 \cite{2016inceptionv3} and ViT \cite{2021ViT}, which are denoted in \protect\colorboxblock{cblue_border}{cblue_fill},  \protect\colorboxblock{corange_border}{corange_fill} and \protect\colorboxblock{cgreen_border}{cgreen_fill}, respectively, and annotate each classifier’s predicted label next to the corresponding viewpoint.
        }\label{fig:waterjug_tar}
	\end{center}
\end{figure*}
\begin{table*}[!t]
\centering
\fontsize{8.1pt}{11pt}\selectfont
\caption{Quantitative comparison of different attack methods under untargeted and targeted attack scenarios. For untargeted attacks, adversarial projections are designed to cause camera-captured instances of the true class \textbf{water jug} to be misclassified as any other class. For targeted attacks, the objective is to consistently fool classifiers into predicting \textbf{coffeepot} for scenes whose true label is \textbf{water jug}. \textbf{OSR} is the success rate for corresponding rows.}
\label{tab:table_compare_waterjug}
\resizebox{\linewidth}{!}{
\begin{tabular}{@{}lcccccccccccccc@{}}
    \toprule[0.5mm]
    \multirow{2}{*}{\textbf{Method}} & \multirow{2}{*}{\textbf{d$_\textbf{thr}$}} 
    & \multicolumn{6}{c}{\textbf{Untargeted attack}} 
    & \multicolumn{6}{c}{\textbf{Targeted attack}} \\ 
    \cmidrule(lr){3-8} \cmidrule(lr){9-14}
     &  & \textbf{$\Delta$E$\downarrow$} & \textbf{L$_\infty\downarrow$} & \textbf{PSNR$\uparrow$} & \textbf{SSIM$\uparrow$} & \textbf{OSR$\uparrow$} & \textbf{Avg. OSR$\uparrow$} 
     & \textbf{$\Delta$E$\downarrow$} & \textbf{L$_\infty\downarrow$} & \textbf{PSNR$\uparrow$} & \textbf{SSIM$\uparrow$} & \textbf{OSR$\uparrow$} & \textbf{Avg. OSR$\uparrow$} \\ 
    \midrule
    \multirow{3}{*}{SPAA \cite{SPAA2022}}
     & 4 & 1.53 & 2.58 & 39.44 & 0.9690 & 51.67\% & \multirow{3}{*}{59.44\%}
         & 1.50 & 2.53 & 39.58 & 0.9703 & 8.33\% & \multirow{3}{*}{6.11\%} \\
     & 5 & 1.69 & 2.99 & 37.89 & 0.9612 & 56.67\% &  
         & 1.73 & 2.98 & 38.01 & 0.9626 & 6.67\% &  \\
     & 6 & 2.04 & 3.49 & 36.61 & 0.9530 & 70.00\% &
         & 2.04 & 3.38 & 37.06 & 0.9571 & 3.33\% &  \\
    \cmidrule[0.3mm]{1-14}
    \multirow{3}{*}{CAPAA \cite{CAPAA2025}}
     & 4 & 1.55 & 2.52 & 39.85 & 0.9750 & 53.33\% & \multirow{3}{*}{56.67\%}
         & 1.49 & 2.45 & 40.17 & 0.9756 & 5.00\% & \multirow{3}{*}{5.56\%} \\
     & 5 & 1.58 & 2.62 & 39.36 & 0.9742 & 58.33\% &
         & 1.43 & 2.33 & 40.80 & 0.9781 & 5.00\% &  \\
     & 6 & 1.58 & 2.65 & 39.21 & 0.9737 & 58.33\% &
         & 1.48 & 2.42 & 40.24 & 0.9760 & 6.67\% &  \\
    \cmidrule[0.3mm]{1-14}
    \multirow{3}{*}{\shortstack[l]{VIPA \\classifier-specific}}
     & 4 & 1.74 & 2.77 & 39.37 & 0.9673 & 61.67\% & \multirow{3}{*}{65.00\%}
         & 1.74 & 2.80 & 39.21 & 0.9662 & 10.00\% & \multirow{3}{*}{11.11\%} \\
     & 5 & 1.74 & 2.77 & 39.39 & 0.9676 & 65.00\% & 
         & 1.74 & 2.79 & 39.24 & 0.9667 & 11.67\% &  \\
     & 6 & 1.75 & 2.79 & 39.32 & 0.9673 & 68.33\% &
         & 1.74 & 2.78 & 39.28 & 0.9666 & 11.67\% &  \\
    \cmidrule[0.3mm]{1-14}
    \multirow{3}{*}{VIPA (ours)}
     & 4 & 1.87 & 3.00 & 38.49 & 0.9632 & 70.00\% & \multirow{3}{*}{\textbf{73.33\%}}
         & 1.82 & 2.96 & 38.63 & 0.9619 & 20.00\% & \multirow{3}{*}{\textbf{17.78\%}} \\
     & 5 & 1.88 & 3.02 & 38.42 & 0.9626 & 75.00\% &
         & 1.81 & 2.93 & 38.69 & 0.9618 & 15.00\% &  \\
     & 6 & 1.87 & 2.98 & 38.54 & 0.9633 & 75.00\% &
         & 1.81 & 2.94 & 38.70 & 0.9620 & 18.33\% &  \\
    \bottomrule[0.5mm]
\end{tabular}
}
\end{table*}

\end{document}